\documentclass[10pt,twocolumn,letterpaper]{article}

\usepackage{cvpr}
\usepackage{times}
\usepackage{epsfig}
\usepackage{graphicx}
\usepackage{amsmath}
\usepackage{amssymb}
\usepackage{array}
\usepackage{subcaption}
\usepackage{multirow}
\usepackage[export]{adjustbox}

\usepackage{float}
\usepackage{placeins}
\restylefloat{table}
\usepackage[pagebackref=true,breaklinks=true,letterpaper=true,colorlinks,bookmarks=false]{hyperref}

\cvprfinalcopy 

\def\cvprPaperID{337} 

\ifcvprfinal\fi

\begin{document}

\title{Going Deeper into First-Person Activity Recognition}
\author{Minghuang Ma, Haoqi Fan and Kris M. Kitani\\
Carnegie Mellon University\\
Pittsburgh, PA 15213, USA\\
\href{mailto:minghuam@andrew.cmu.edu}{\tt\small minghuam@andrew.cmu.edu} 
\href{mailto:haoqif@andrew.cmu.edu}{\tt\small haoqif@andrew.cmu.edu} 
\href{mailto:kkitani@cs.cmu.edu}{\tt\small kkitani@cs.cmu.edu} 
}

\maketitle

\begin{abstract}
We bring together ideas from recent work on feature design for egocentric action recognition under one framework by exploring the use of deep convolutional neural networks (CNN). Recent work has shown that features such as hand appearance, object attributes, local hand motion and camera ego-motion are important for characterizing first-person actions. To integrate these ideas under one framework, we propose a twin stream network architecture, where one stream analyzes appearance information and the other stream analyzes motion information. Our appearance stream encodes prior knowledge of the egocentric paradigm by explicitly training the network to segment hands and localize objects. By visualizing certain neuron activation of our network, we show that our proposed architecture naturally learns features that capture object attributes and hand-object configurations. Our extensive experiments on benchmark egocentric action datasets show that our deep architecture enables recognition rates that significantly outperform state-of-the-art techniques -- an average $6.6\%$ increase in accuracy over all datasets. Furthermore, by learning to recognize objects, actions and activities jointly, the performance of individual recognition tasks also increase by $30\%$ (actions) and $14\%$ (objects). We also include the results of extensive ablative analysis to highlight the importance of network design decisions.
\end{abstract}

\vspace*{-4mm}
\section{Introduction}
\vspace*{-1mm}

Recently there has been a renewed interest in the use of first-person point-of-view cameras to better understand human activity. In order to accurately recognize first-person activities, recent work in first-person activity understanding has highlighted the importance of taking into consideration both \textit{appearance} and \textit{motion} information. Since the majority of actions are centered around hand-object interactions in the first-person sensing scenario, it is important to capture \textit{appearance} corresponding to such features as hand regions, grasp shape, object type or object attributes. Capturing \textit{motion} information such as local hand movements and global head motion, is another important visual cue as the temporal motion signature can be used to differentiate between complementary actions such as \textit{take} and \textit{put} or periodic actions such as the \textit{cut with knife} action. It is also critical to reason about appearance and motion \textit{jointly}. It has been shown in both third-person \cite{gupta2009observing} and first-person activity analysis \cite{li2015delving} that these two streams of activity information, appearance and motion, should be analyzed jointly to obtain best performance.

\begin{figure}[t]
\centering
\includegraphics[width=0.9\linewidth]{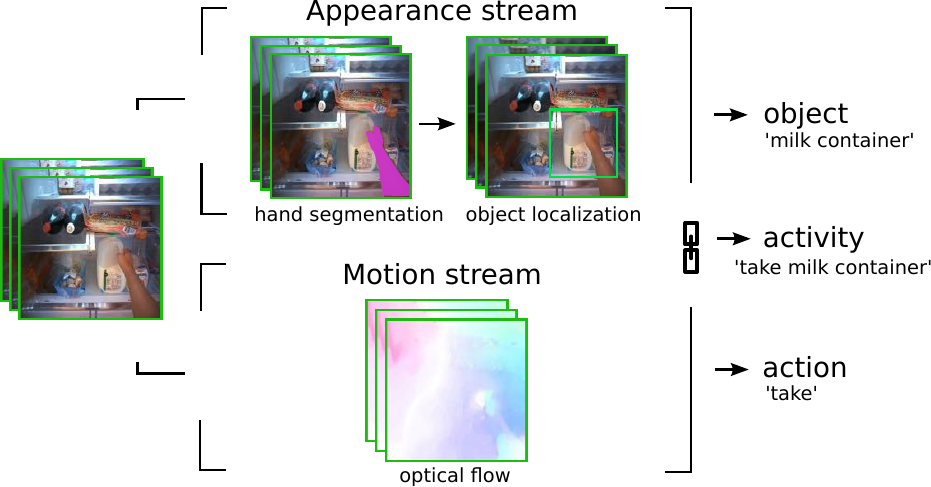}
\caption{Approach overview. Our framework integrates both appearance and motion information. The appearance stream captures hand configurations and object attributes to recognize objects. The motion stream captures objects motion and head movement to recognize actions. The two streams are also learned jointly to recognize activities.}
\label{fig-overview}
\vspace{-6mm}
\end{figure}

Based on these insights, we propose a deep learning architecture designed specifically for egocentric video, that integrates both action appearance and motion within a single model \ref{fig-overview}. More specifically, our proposed network has a two stream architecture composed of an appearance-based CNN that works on localized object of interest image frames and a motion-based CNN that uses stacked optical flow fields as input. Using the terminology of \cite{fathi2011understanding}, we use late fusion with a fully-connected top layer to formulate a multi-task prediction network over \textit{actions}, \textit{objects} and \textit{activities}. The term \textit{action} describes motions such as \textit{put}, \textit{scoop} or \textit{spread}. The term \textit{object} refers to item such as bread, spoon or cup. The term \textit{activity} is used to represent an action-object pair such as \textit{take milk container}. 

The appearance-based stream is customized for ego-centric video analysis by explicitly training a hand segmentation network to enable an attention-based mechanism to focus on certain regions of the image near the hand. The appearance-based stream is also trained with object images cropped based on hand location to identify objects of manipulation. In this way, the appearance-based stream is enabled to encode such features such as hand-object configurations and object attributes.

The motion-based stream is a generalized CNN that takes as input a stack of optical-flow motion fields. This stream is trained to differentiate between action labels such as \textit{put}, \textit{take}, \textit{close}, \textit{scoop} and \textit{spread}. Instead of compensating for camera ego-motion as a pre-processing step, we allow the network to automatically discover which motion patterns (camera, object or hand motion) are most useful for discriminating between action types. Results show that the network automatically learns to differentiate between different motion types.

We train the appearance stream and motion stream jointly as a multi-task learning problem. Our experiments show that by learning the parameters of our proposed network jointly, we are able to outperform state-of-the-art techniques by over $6.6\%$ on the task of egocentric activity recognition without the use of gaze information, and in addition improve the accuracy of each sub-task (30\% for action recognition and 14\% object recognition). 

Perhaps more importantly, the trained network also helps to better understand and to reconfirm the value of key features needed to discriminate between various egocentric activities. We include visualizations of neuron activations and show that the network has learned intuitive features such as hand-object configurations, object attributes and hand motion signatures isolated from global motion.

\textbf{Contributions}: (1) we formulate a deep learning architecture customized for ego-centric vision; (2) we obtain state-of-the-art performance propelling the field towards higher performance; (3) we provide ablative analysis of design choices to help understand how each component contributes to performance; and (4) visualization and analysis of the resulting network to understand what is being learned at the intermediate layers of the network. The related work is summarized as follows.

\noindent \textbf{Human Activity Recognition:} Traditionally, in video-based human activity understanding research \cite{aggarwal2011human,peng2014bag}, many approaches make use of local visual features like HOG \cite{laptev2008learning}, HOF \cite{laptev2008learning} and MBH \cite{wang2013dense} to encode appearance information. These features are typically extracted from spatio-temporal keypoints \cite{laptev2005space} but can also be extracted over dense trajectories \cite{wang2011action, wang2013action}, which can improve recognition performance. Most recently, it has been shown that the visual feature representation can be learned automatically using a deep convolutional neural network for image understanding tasks \cite{krizhevsky2012imagenet}. In the realm of action recognition, Simonyan and Zisserman \cite{simonyan2014two} proposed a two-stream network to capture spatial appearance on still images and temporal motion between frames. Ji \etal \cite{ji20133d} used 3D convolutions to extract both spatial and temporal features using a one stream network. Wang \etal \cite{wang2015action} further develops trajectory-pooled deep-convolutional descriptor (TDD) to incorporate both specially designed features and deep-learned features to achieve state-of-the-art results.

\noindent \textbf{First-Person Video Analysis: }  In a similar fashion to third-person activity analysis, the first-person vision community has also explored various types of visual features for representing human activity. Kitani \etal \cite{kitani2011fast} used optical flow-based global motion descriptors to discover ego-action in sports videos. Spriggs \etal \cite{spriggs2009temporal} performed activity segmentation GIST descriptors. Pirsiavash \etal \cite{pirsiavash2012detecting} developed a composition of HOG features to model object and hand appearance during an activity. Bambach \etal \cite{bambach2015lending} used hand regions to understand activity. Fathi \etal proposed mid-level motion features and gaze for recognizing ego-centric activities in \cite{fathi2012learning, fathi2008action}. To encode first-person videos using those features, the most prevalent representations are BoW and improved Fisher Vector \cite{perronnin2010improving}. In \cite{li2015delving}, Li \etal performed a systemic evaluation of features and provided a list of best practices of combining different cues to achieve state-of-the-art results for activity recognition. Similar to third-person vision activity recognition research, there has also been a number of attempts to use CNN for understanding activities in first-person videos. Ryoo \etal \cite{ryoo2014pooled} develops a new pooled feature representation and shows superior performance using CNN as a appearance feature extractor. Poleg \etal \cite{poleg2015compact} proposes to use temporal convolutions over optical flow motion fields to index first-person videos. However, a framework to integrate the success of ego-centric features and the power of CNNs is still missing due to challenges of feature diversity and limited training data. In this paper, we aim to design such a framework to address the problem of ego-centric activity recognition.

\vspace*{-1mm}
\section{Egocentric Activity Deep Network}
\vspace*{-1mm}

\begin{figure*}[t]
\vspace{-2mm}
\centering
\includegraphics[width=0.9\linewidth]{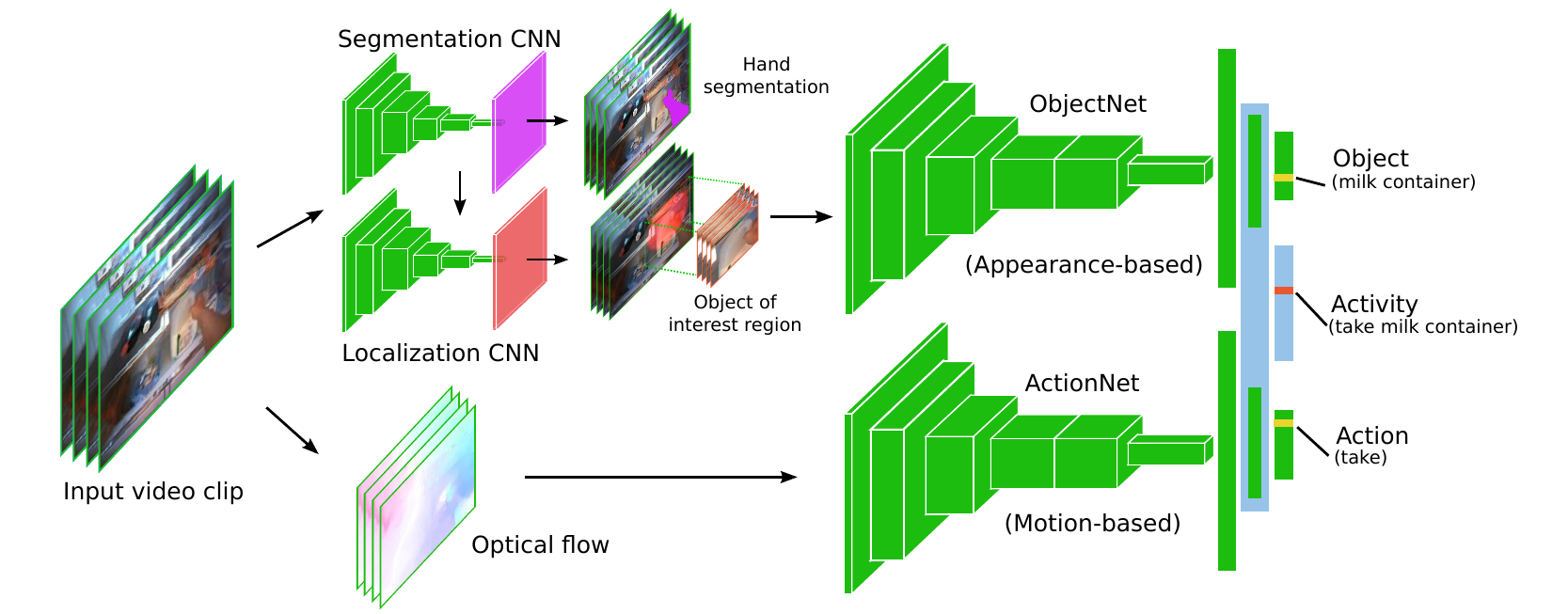}
\caption{Framework architecture for action, object and activity recognition. Hand segmentation network is first trained to capture hand appearance. It is then fine-tuned to a localization network to localize object of interest. Object CNN and motion CNN are then trained separately to recognize objects and actions. Finally, the two networks are fine-tuned jointly with a triplet loss function to recognize objects, actions and activities. This proposed network beats all baseline models.}
\vspace{-6mm}
\label{fig-architecture}
\end{figure*}

We describe our proposed deep network architecture for recognizing activity labels from short video clips taken by an egocentric camera. As we have argued above, the manifestation of an activity can be decomposed into observed appearance (hand and objects) and observed motion (local hand movement and user ego-motion). Based on this decomposition, we develop two base networks: (1) ObjectNet takes a single image as input to determine the \textit{appearance} features of the activity and is trained using \textit{object} labels; (2) ActionNet takes a sequence of optical flow fields to determine the \textit{motion} features of the activity and is trained using \textit{action} labels. Taking the output of both of these networks, we use a late fusion step to concatenate the output of the two networks and uses the joint representation to predict three outputs, namely, \textit{action}, \textit{object} and \textit{activity}. More formally, given a short video sequence of $N$ image frames $\mathbf{I} = \{I_1,\dots, I_N\}$, our network predicts three output labels: $\{y_{\textrm{object}}, y_{\textrm{action}}, y_{\textrm{activity}}\}$. The architecture of the entire network is illustrated in Figure \ref{fig-architecture}.

\vspace*{-1mm}
\subsection{ObjectNet: Recognizing Objects from Appearance}\label{learning_object}
\vspace*{-1mm}

As shown in \cite{ghosh2012discovering, pirsiavash2012detecting}, recognizing objects in videos is an important aspect of ego-centric activity understanding. We aim to predict the object label $y_{\textrm{object}}$ in this section. To do so, we are particularly interested in the object being interacted with or manipulated -- the \textit{object of interest}. However, detecting all objects accurately in the scene is difficult. It also provides limited information about the interested object. Our proposed model will first localize and then recognize the object of interest.

\begin{figure}[t]
\centering
\includegraphics[width=0.9\linewidth]{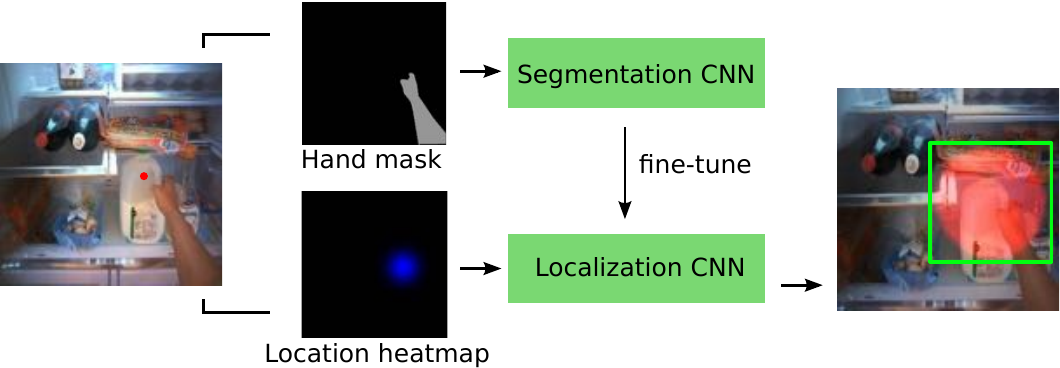}
\caption{Pipeline for localization network training. Hand segmentation network is first trained using images and binary hand masks. Localization network is then fine-tuned from hand segmentation network using images and object location heatmaps synthesized from object locations.}
\label{fig-obj-arch}
\vspace*{-6mm}
\end{figure}

Although we can assume that the object of interest is often located at the center of the subject's reachable region, it is not always present at the center of the camera image due to head motion. Instead, we observe that the object of interest most frequently appears in the vicinity of hands. A similar observation was also made in \cite{li2013learning, fathi2011learning}. Besides hand location, hand pose and shape is also important to estimate the manipulation points as shown in \cite{li2013learning}. We therefore seek to segment the hands out of the image and use hand appearance to predict the location of the object of interest. We first train a pixel-to-pixel hand segmentation network using raw images and binary hand masks. This network will output a hand probability map. To predict object of interest location using this hand representation, a naive approach is to build a regression model on top. For instance, we can train another CNN or a regressor using features from the hand segmentation network. However, our experiments with this approach achieve low performance due to limited training data. The prediction tends to favor the image center as it is where the object of interest occurs most frequently. Our final pipeline is illustrated in Figure \ref{fig-obj-arch}. After training the hand segmentation network, we fine-tune a localization network to predict a pixel-level object occurrence probability map. Inspired by previous work in pose estimation \cite{pfister2015flowing}, we synthesize a heatmap by placing a 2D Gaussian distribution at the location of the object of interest. We use this heatmap as ground-truth and use per-pixel Euclidean loss to train the network. To transfer the hand representation learned from the segmentation network, we initialize the localization network with the weights from the segmentation network and then fine-tune the localization network with the new loss layer. The details of the segmentation and localization CNNs are listed as follows.

\noindent \textbf{(1) Hand segmentation network: } For training data, we can either use annotated ground-truth hand masks or output of pixel-level hand detectors like \cite{li2013pixel}. For the network architecture, we use a low resolution FCN32-s as in \cite{long2014fully} as it is a relatively smaller model and converges faster. The loss function for the segmentation network is the sum of per-pixel two-class softmax losses.

\def\LW{\dimexpr.8\linewidth-.5em}
\def\LWW{\dimexpr.075\linewidth-.5em}
\begin{figure}
\centering {}
\parbox{\LWW}{(a)}
\parbox{\LW}{\includegraphics[width=0.8\linewidth]{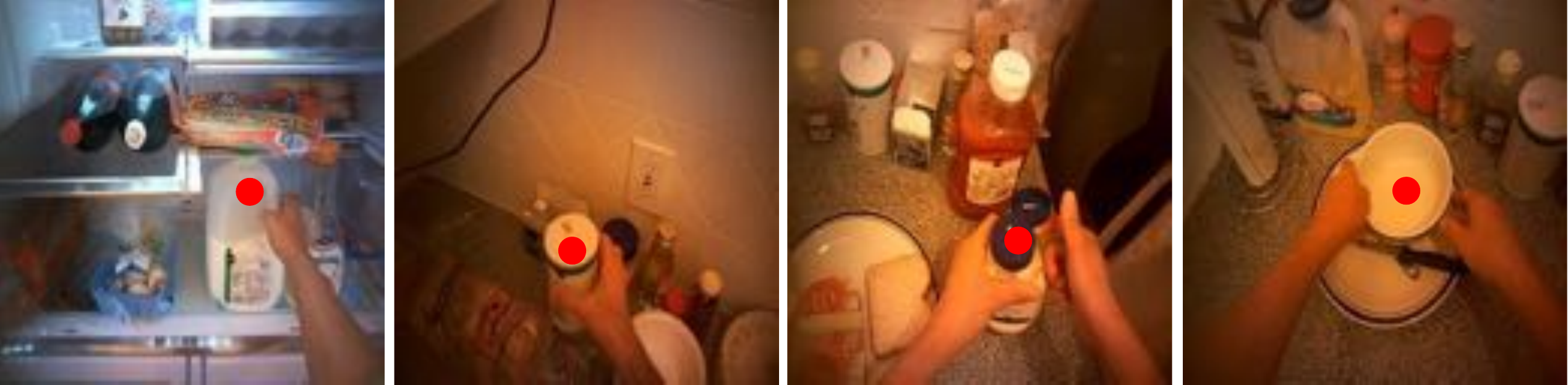}}\\
\vspace{1pt}
\parbox{\LWW}{(b)}
\parbox{\LW}{\includegraphics[width=0.8\linewidth]{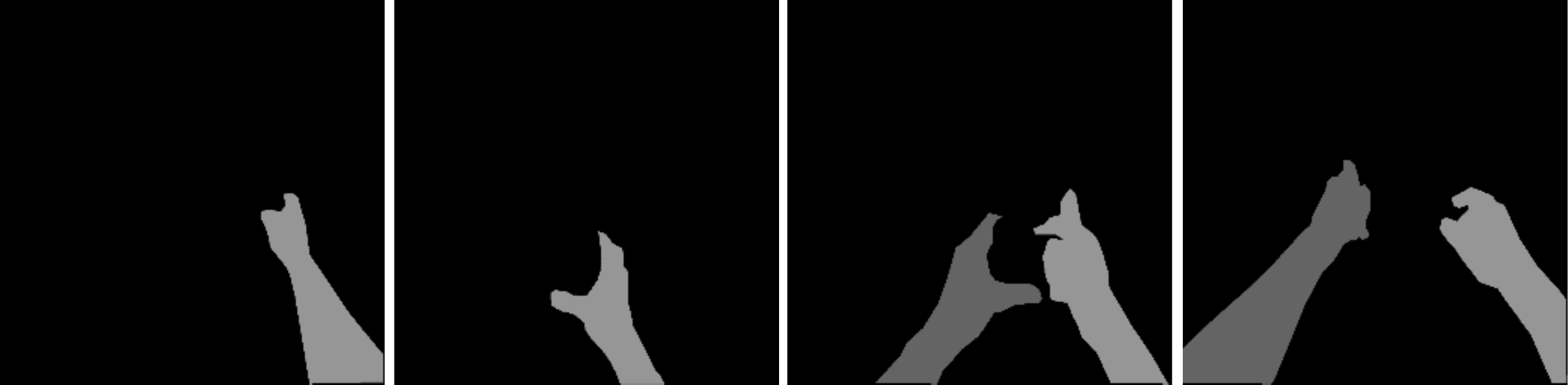}}\\
\vspace{1pt}
\parbox{\LWW}{(c)}
\parbox{\LW}{\includegraphics[width=0.8\linewidth]{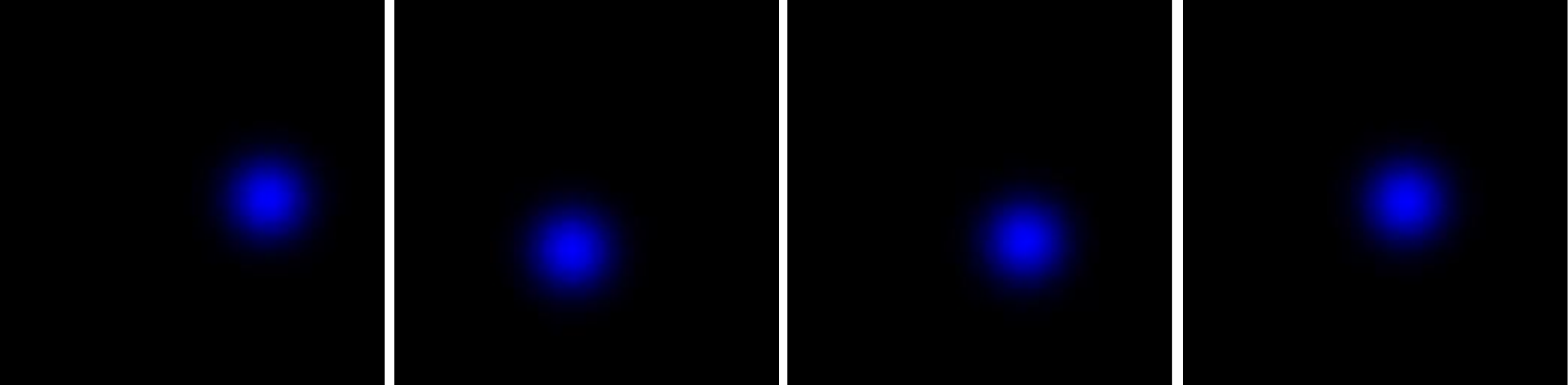}}\\
\vspace{-2mm}
\caption{Training data examples for localization CNN. (a) Raw video images with annotated object locations (in red). (b) Ground-truth hand masks which can be annotated manually or generated using hand detectors such as \cite{li2013pixel}. (c) Synthesized location heat-maps by placing a Gaussian bump at the object location.}
\label{fig-obj-data}
\vspace{-6mm}
\end{figure}

\noindent \textbf{(2) Object localization network: } For training data, we first manually annotate object of interest locations in the training images of the hand segmentation network. We then synthesize the location heatmaps using a Gaussian distribution as discussed above. Examples of training data are shown in Figure \ref{fig-obj-data}. We use the same FCN32-s network architecture and replace the softmax layer with a per-pixel Euclidean loss layer.

To this extent, we have trained an object localization network that outputs a per-pixel occurrence probability map of the object of interest. To generate the final object region proposals, we first run the localization network on input image sequence $\mathbf{I} = \{I_1, I_2, \dots, I_N\}$ and generate probability maps of object locations $\mathbf{H} = \{H_1, H_2,\dots, H_N\}$. We then threshold each probability map and use the centroid of the largest blob as the predicted center of the object. We then crop the object out of the raw image at the predicted center using a fixed-size bounding box. We fix the crop size and ignore the scale difference by observing that the object of interest is always within the reachable distance of the subject. In this way, we generate a sequence of cropped object region images $\mathbf{O} = \{O_1, O_2, \dots, O_N\}$ as the input of the object recognition CNN. The localization result is stable on a per-frame basis, hence there is no temporal smoothness adopted. 

With the cropped image sequence of objects of interest $\mathbf{\{O_i\}}$, we then train the object CNN using the model of CNN-M-2048 \cite{chatfield2014return} to recognize the objects. We choose this network architecture due to its high performance on ImageNet image classification. Since a better architecture of base network is out of the scope of this work, we use this network as our base network in this paper unless otherwise mentioned. We adapt it to different tasks (e.g. action recognition) with minimum modifications in this paper. For object recognition, we train the network using $\{(O_i, y_{\text{object}}\})$ pairs as training data and softmax as the loss function. At testing time, we run the network on the cropped object image $O_i$ to predict object class scores. We then calculate the mean score of all frames in a sequence for each activity class and select the activity label with largest mean score as the final predicted label of object. 

Up until now, we have trained a localization network to localize the object of interest by explicitly incorporating hand appearance. Using the cropped images of the localized object of interested, we have trained an object recognition network to predict the object label $y_{\textrm{object}}$. We will show later that this recognition pipeline also captures important appearance cues such as object attributes and hand appearance. We now move forward to the motion stream of our framework.

\vspace*{-1mm}
\subsection{ActionNet: Recognizing Actions from Motion} \label{learning_action}
\vspace*{-1mm}

In this section, our target is to predict the action label $y_{\textrm{action}}$ from motion. Unlike straightforward appearance cues like hands and objects discussed in previous section, motion features in ego-centric videos are more complex because the head motion might cancel the object and hand motion. Although Wang \etal \cite{wang2013action} shows that compensation of camera motion improves accuracy in traditional action recognition tasks, for ego-centric videos, background motion is often a good estimation of head motion and thus an important cue for recognizing actions \cite{li2015delving}. Instead of decoupling foreground (object and hand) motion and background (camera) motion and calculating features separately, we aim to use CNN to capture different local motion features and temporal features together implicitly.

In order to train a CNN network with motion input, we follow \cite{simonyan2014two} to use optical flow images to represent motion information. In particular, given a video sequence of $N$ frames $\mathbf{I} = \{I_1, I_2, ..., I_N\}$ and corresponding action label $y_{\textit{\textrm{action}}}$, we first calculate optical flow of each two consecutive frames and encode the horizontal and vertical flow separately in $\mathbf{U} = \{U_1, U_2, ..., U_{N-1}\}$ and $\mathbf{V} = \{V_1, V_2, ..., V_{N-1}\}$. To incorporate temporal information, we use a fixed length of $L$ frames and stack corresponding optical flow images together as input samples of the network noted as $\mathbf{X} = \{X_1,...,X_{N-L+1}\}$ where $X_i = \{U_i, V_i,..., U_{i+L-1}, V_{i+L-1}\}$. 

With motion represented in optical flow images, we train the motion CNN using $\{(X_i, y_{\textit{\textrm{action}}})\}$ pairs as training data and softmax as the loss function. At testing time, we run the network on input motion data ${X_i}$ to predict the scores for each action class. We then average the scores for all frames in the sequence and pick the action class with maximum average score as the predicted label of the action. With the learned representation of objects and actions, we now move to the next section for activity recognition.

\vspace*{-1mm}
\subsection{Fusion Layer: Recognizing Activity} \label{sec-joint-training}
\vspace*{-1mm}

In this section, we seek to recognize the activity label $y_{\textrm{activity}}$ given the representations learned from the two network streams in previous sections. A natural approach is to use the two networks as feature extractors and training a classifier using activity labels. However, this approach ignores the co-relation between actions, objects and activities. For instance, if we are confident that the action is \textit{stir} from repeated circular motion, it is highly probable that the object is \textit{tea} or \textit{coffee}. In the other way, if we know the object is \textit{tea} or \textit{coffee}, the probability that the action is \textit{cut} or \textit{fold} should be very low. Based on this intuition, we fuse the action and object networks together as one network by concatenating the second last fully connected layers from the two networks and add another fully connected layer on top. We then add another loss layer for activity on top. The final fused network therefore has three weighted losses: action, object and activity loss. Then weighted sum of three losses is calculated as the overall loss. We can set the weights empirically by the relative importance of three tasks and train one network to learn activity, action and object simultaneously. The loss function for the final network can be formulated as $L_{\textrm{network}} = w_1\cdot L_{\textrm{action}} + w_2\cdot L_{\textrm{object}} + w_3\cdot L_{\textrm{activity}}$.

To train the fused network, we transfer the weights of the trained motion CNN and object CNN and fine-tune it to recognize the activity. Specifically, given a video sequence of $N$ frames $\mathbf{I} = \{I_1, I_2, ..., I_N\}$, we follow section \ref{learning_object} to localize the objects of interest and get a sequence of object images $\mathbf{O} = \{O_1, O_2, ..., O_N\}$. We follow section \ref{learning_action} to calculate optical flow image pairs $\{\mathbf{U}, \mathbf{V}\}$ and stack them using a fixed length of $L$ frames into $\mathbf{X} = \{X_1,...,X_{N-L+1}\}$ where $X_i=(U_i, V_i)$. At training time, for each optical flow blob $X_i$, we randomly pick a object image $O_{j}$ where $i \leq j < i+L$ and form the training data pair $(X_i, O_j, y_{\textrm{action}}, y_{\textrm{object}}, y_{\textrm{activity}})$. This is also a way to augment the training data to avoid over-fitting. At testing time, we pick the center object image frame such that $j = (2i+L)/2$ as the annotated boundary of an activity sequence is loose. We run the network on all data pairs to predict the scores for activity. We then average the scores and use the activity class with maximum average score as the predicted activity label.

\vspace*{-2mm}
\section{Experiments}
\vspace*{-1mm}

We briefly introduce the datasets in Section \ref{sec-dataset} and describe the details for training networks in Section \ref{sec-net-training}. We then present experimental results for the three tasks of object recognition (Section \ref{sec-object-results}), action recognition (Section \ref{sec-action-results}) and activity recognition (Section \ref{sec-activity-results}).

\vspace*{-1mm}
\subsection{Dataset}\label{sec-dataset}
\vspace*{-1mm}

We run experiments on three public datasets: GTEA, GTEA gaze (Gaze) and GTEA gaze+ (Gaze+) as these datasets were collected using a head-mount camera and most of the activities involve hand-object interactions. The annotation label for each activity contains a verb (action) and a set of nouns (object). We perform all our experiments using leave-one-subject-out cross-validation. We also report results on fixed-splits following previous work.

\noindent \textbf{GTEA:} This dataset \cite{fathi2011learning} contains 7 types of activities performed by 4 different subjects. There are 71 activity categories and 525 instances in the original labels. We report comparative results on two subsets used in previous work \cite{fathi2013modeling, li2015delving, fathi2011learning}: 71 classes and 61 classes. \textbf{Gaze:} This dataset \cite{fathi2012learning} contains 17 sequences performed by 14 different subjects. There are 40 activity categories and 331 instances in the original labels. We report results on two subsets used in previous works \cite{li2015delving, fathi2012learning}: 40 classes and 25 classes. \textbf{Gaze+:} This dataset \cite{fathi2012learning} contains 7 types of activities performed by 10 different subjects. We report results on a 44 classes subset with 1958 action instances following \cite{li2015delving}.

\vspace*{-1mm}
\subsection{Network Training}\label{sec-net-training}
\vspace*{-1mm}
For network architecture, we use FCN32-s \cite{long2014fully} for hand segmentation and object localization. We use CNN-M-2048 \cite{chatfield2014return} for action and object recognition. Due to the limited sizes of the three public datasets, we adopt the fine-tuning \cite{oquab2014learning} approach to initialize our networks. Specifically, we use available pre-trained models from three large-scale datasets: UCF101\cite{soomro2012ucf101}, Pascal-Context\cite{mottaghi2014role} and ImageNet\cite{deng2009imagenet} for motion, hand segmentation and object CNN respectively.

\textbf{Data augmentation}. To further address the problem of limited data, we apply data augmentation \cite{krizhevsky2012imagenet} to improve generalization of CNN networks. \textbf{Crop}: All of our network inputs are resized to $K\times C\times 256\times 256$, where $K$ is batch size, $C$ is input channels. We randomly crop them to $K\times C\times 224\times 224$ at training time. \textbf{Flip}: We randomly mirror input images horizontally. For optical flow frames $(U_i, V_i)$, we mirror them to $(255 - U_i, V_i)$. \textbf{Replication}: We also replicate training data by repeating minority classes to match with majority classes at training time.

\textbf{Training details}. We use a modified version of Caffe \cite{jia2014caffe} and Nvidia Titan X GPU to train our networks. We use stochastic gradient descent with momentum as our optimization method. We use a fixed learning rate of $\gamma=1e-8$ for fine-tuning hand segmentation and object localization CNNs, $\gamma=5e-4$ for motion CNN and $\gamma=1e-4$ for object CNN. For joint training, we lower the learning rate of two sub-networks by a factor of 10. We use batch sizes of 16, 128, 180 for localization, object and motion CNNs respectively.

\vspace*{-1mm}
\subsection{ObjectNet Performance}\label{sec-object-results}
\vspace*{-1mm}

We evaluate the localization network and object recognition network of the ObjectNet stream.

\noindent \textbf{Localizing object of interest.} As described in Section \ref{learning_object} and illustrated in Figure \ref{fig-obj-arch}, we first train a hand segmentation to learn the bottom layers of the object localization network. The intuition behind this training procedure is to purposefully bias the object localization network to use the hands as evidence to infer an object bounding box. We first train a hand segmentation network using the model of \cite{long2014fully} to capture hand appearance information. We then swap out the top layer for segmentation with a top layer optimized for object localization (\ie, fine-tune the network to repurpose it for object localization). The network has an input size of $K\times 3\times 256\times 256$ where $K$ is the batch size. After five convolutional ($conv1-conv5$) layers of $2\times 2$ pooling operations, the input image dimension is down-sampled to $1/32$ of the original size. The final deconvolution layer up-samples it back to the original size of $K\times 2\times 256\times 256$. We use raw images and hand masks provided with GTEA and Gaze as training data for the hand segmentation network. Since Gaze+ is not annotated with hand masks, we use \cite{li2013pixel} to detect hands and use the result to train the network. Once the segmentation network is trained, we use manually annotated training images of object locations to re-purpose the the network for localization. Instead of using raw object locations (exact center position of the object), we place a Gaussian bump at the center position to create a heat-map representation as described in Section \ref{learning_object}. 

Figure \ref{fig-local-results} shows qualitative results of the localization network. The localization network successfully predicts the key object of interest out of other irrelevant objects in the scene. Notice that the result is strongly tied to the hand as the network is \textit{pre-train} for hand segmentation. The results also show that the model can deal seamlessly with different hand configurations like one-hand or two-hand scenarios.

\def\LW{\dimexpr.9\linewidth-.5em}
\def\LWW{\dimexpr.075\linewidth-.5em}
\begin{figure}
\vspace{-4mm}
\centering {}
\parbox{\LWW}{(a)}
\parbox{\LW}{\includegraphics[width=0.9\linewidth]{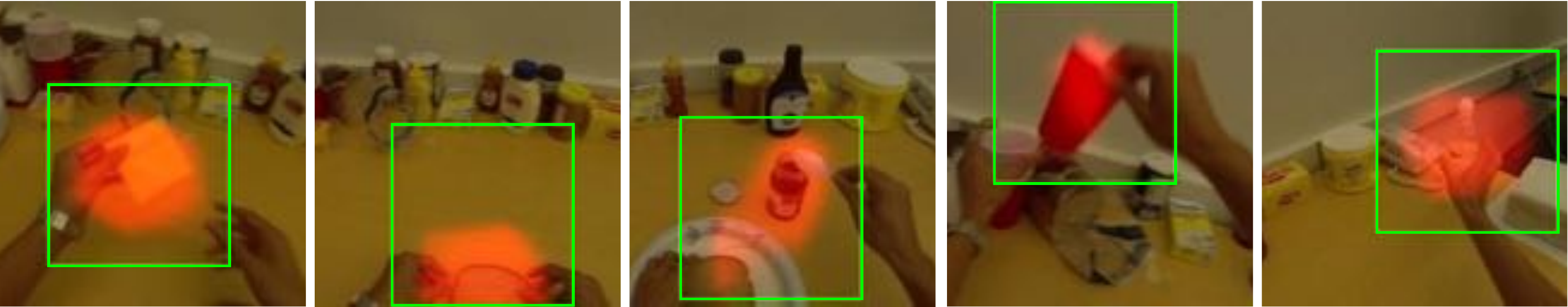}}\\
\vspace{1pt}
\parbox{\LWW}{(b)}
\parbox{\LW}{\includegraphics[width=0.9\linewidth]{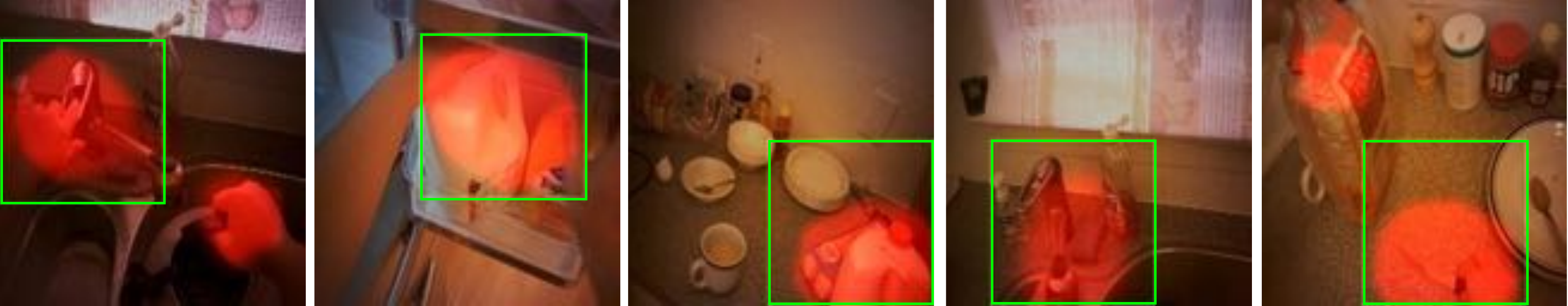}}\\
\vspace{-3pt}
\caption{
Object localization using Hand Information. Visualization of object location probability map (\textcolor{red}{\textbf{red}}) and object bounding box (\textcolor{green}{\textbf{green}}). (a: GTEA, b: Gaze+)
}
\label{fig-local-results}
\vspace{-8mm}
\end{figure}

\noindent \textbf{Recognizing object of interest.} The localized object images are used to train the object CNN. Table \ref{tab-obj-results} compares the performance of our proposed methods with \cite{fathi2011understanding}. Our proposed method dramatically outperforms \cite{fathi2011understanding} by $14\%$. As seen in Table \ref{tab-obj-results} the boost in performance can be attributed to improved localization through the use of hand segmentation-based pre-training.

\begin{table}[t]
\renewcommand{\arraystretch}{1.05}
\centering
\footnotesize
\begin{tabular}{l|c|c|c}
\hline
Object Recognition	    &GTEA(71)	    &Gaze(40)     &Gaze+(44)\\
\hline
Fathi \textit{et al.}\ \cite{fathi2011learning}  & 61.36    & N/A  & N/A\\
\hline
Object CNN              & 67.74             & 38.05          & 61.87\\
\hline
Joint training (Ours)          & \textbf{76.15}          & \textbf{55.55}          & \textbf{74.34}\\
\hline
\end{tabular}
\caption{
Average object recognition accuracy. Proposed method performs 14\% better than the baseline. Joint training of motion and object networks improves accuracy across all datasets.}
\label{tab-obj-results}
\vspace{-2mm}
\end{table}

\def\LW{\dimexpr.9\linewidth-.5em}
\def\LWW{\dimexpr.075\linewidth-.5em}
\begin{figure}
\vspace{-4mm}
\centering {}
\parbox{\LWW}{(a)}
\parbox{\LW}{\includegraphics[width=0.9\linewidth]{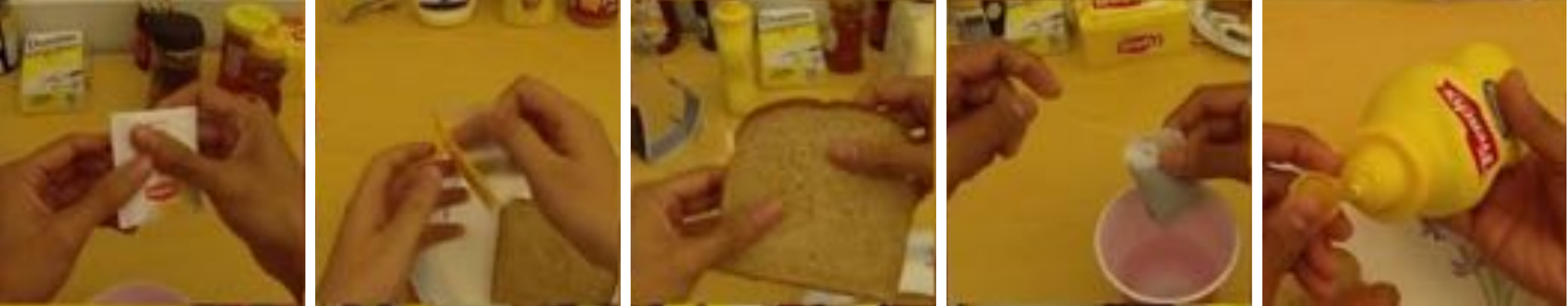}}\\
\vspace{3pt}
\parbox{\LWW}{(b)}
\parbox{\LW}{\includegraphics[width=0.9\linewidth]{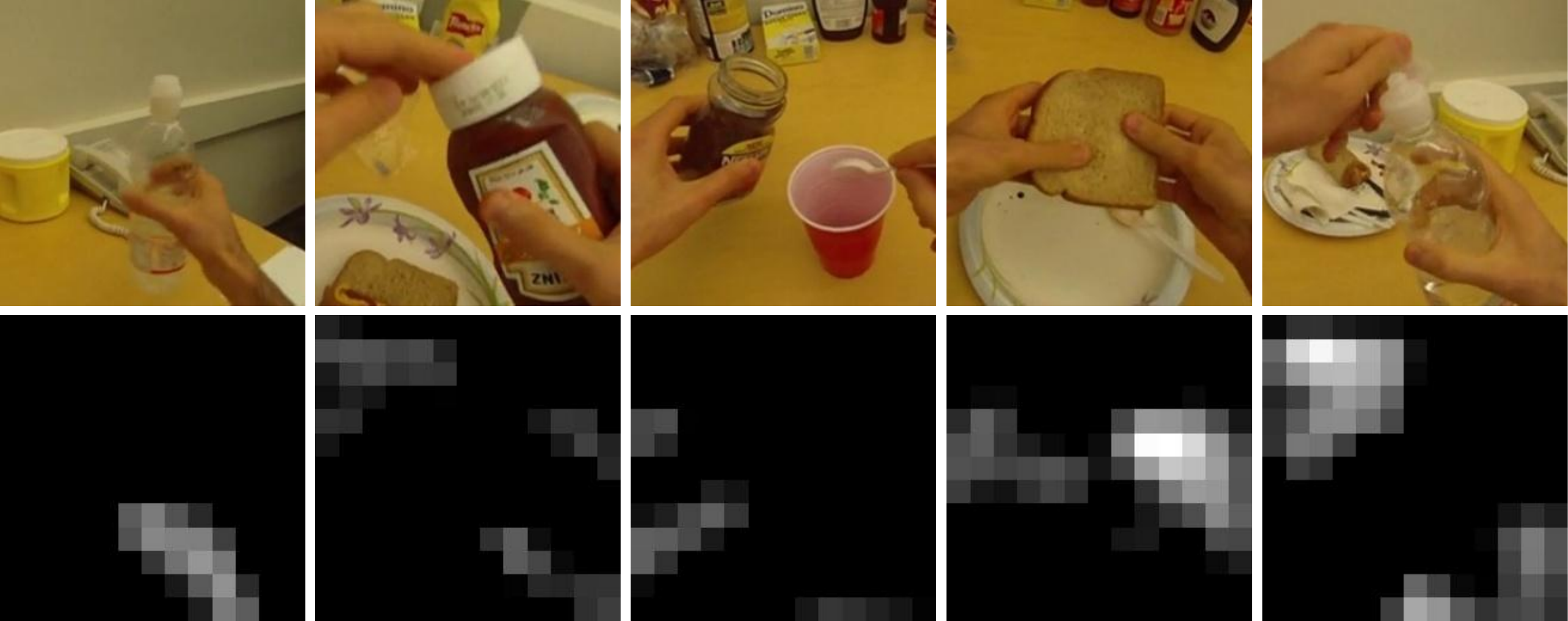}}\\
\vspace{1pt}
\caption{(a) Top 5 training images with strongest activations from the $50^{\text{th}}$ neuron unit in the $conv5$ layer. (b) 5 test images (top row) and $13\times 13$ activations (bottom row) of the same unit. The visualization shows that this unit responds strongly to hand regions. The object network is capturing hand appearance.}
\label{fig-hand-activations}
\vspace{-5mm}
\end{figure}

\begin{figure}[t]
\centering
\footnotesize
  \begin{subfigure}[b]{0.11\textwidth}
    \includegraphics[width=\textwidth]{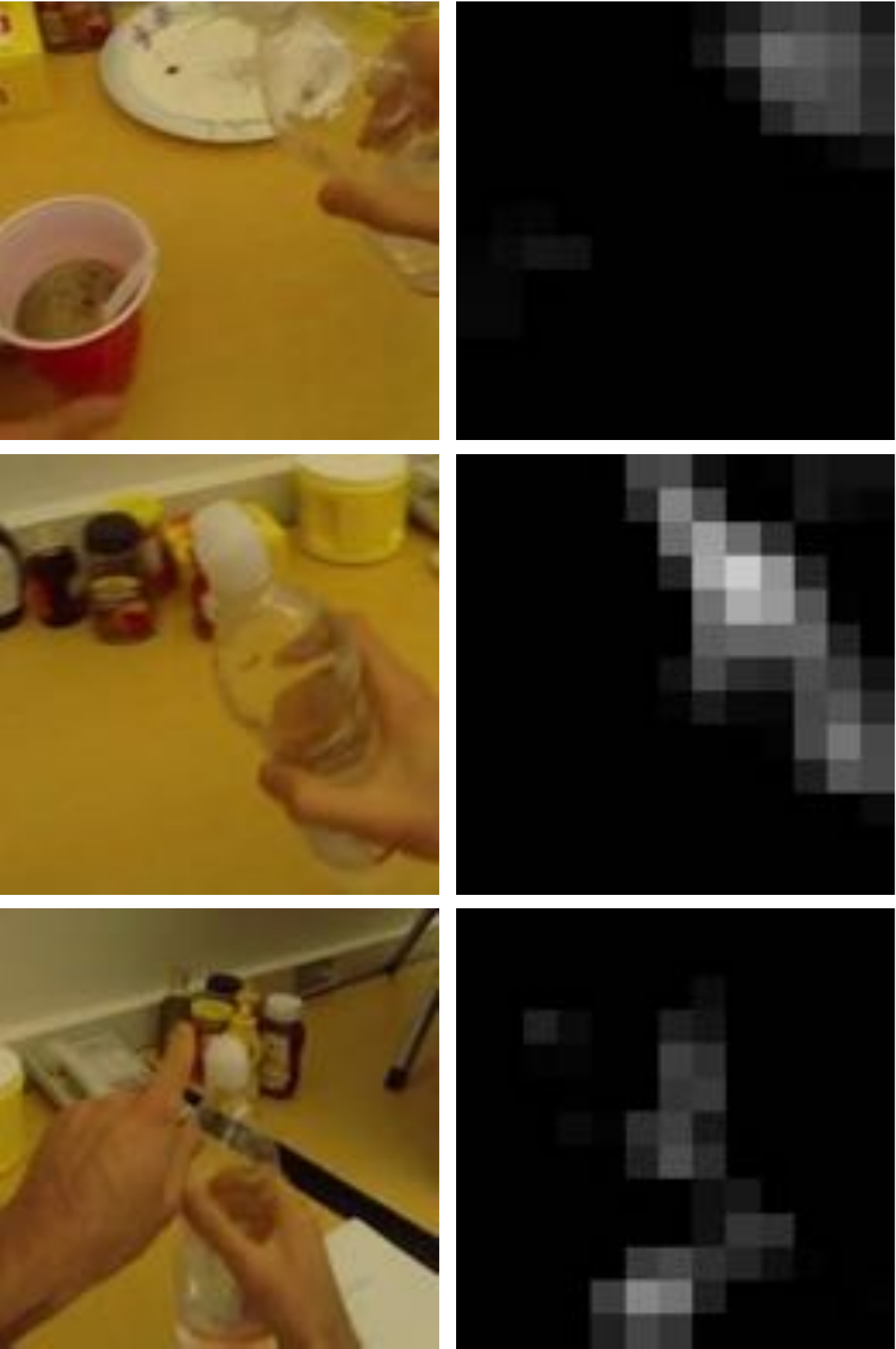}
    \caption{}
    \label{fig:f1}
    \vspace{-2mm}
  \end{subfigure}
  \hspace{2pt}
  \begin{subfigure}[b]{0.11\textwidth}
    \includegraphics[width=\textwidth]{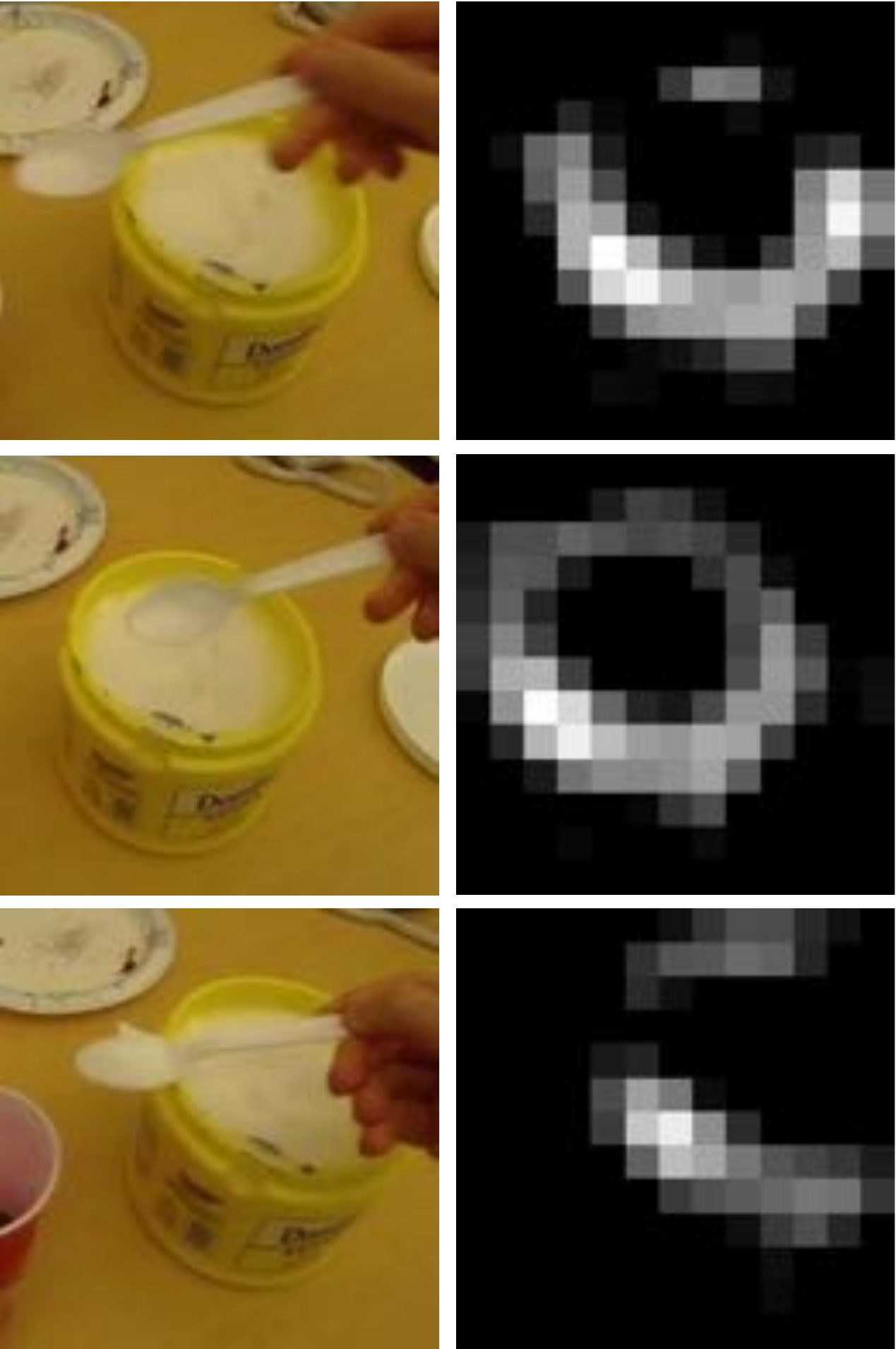}
    \caption{}
    \label{fig-obj-activations-b}
    \vspace{-2mm}
  \end{subfigure}
  \hspace{2pt}
  \begin{subfigure}[b]{0.11\textwidth}
    \includegraphics[width=\textwidth]{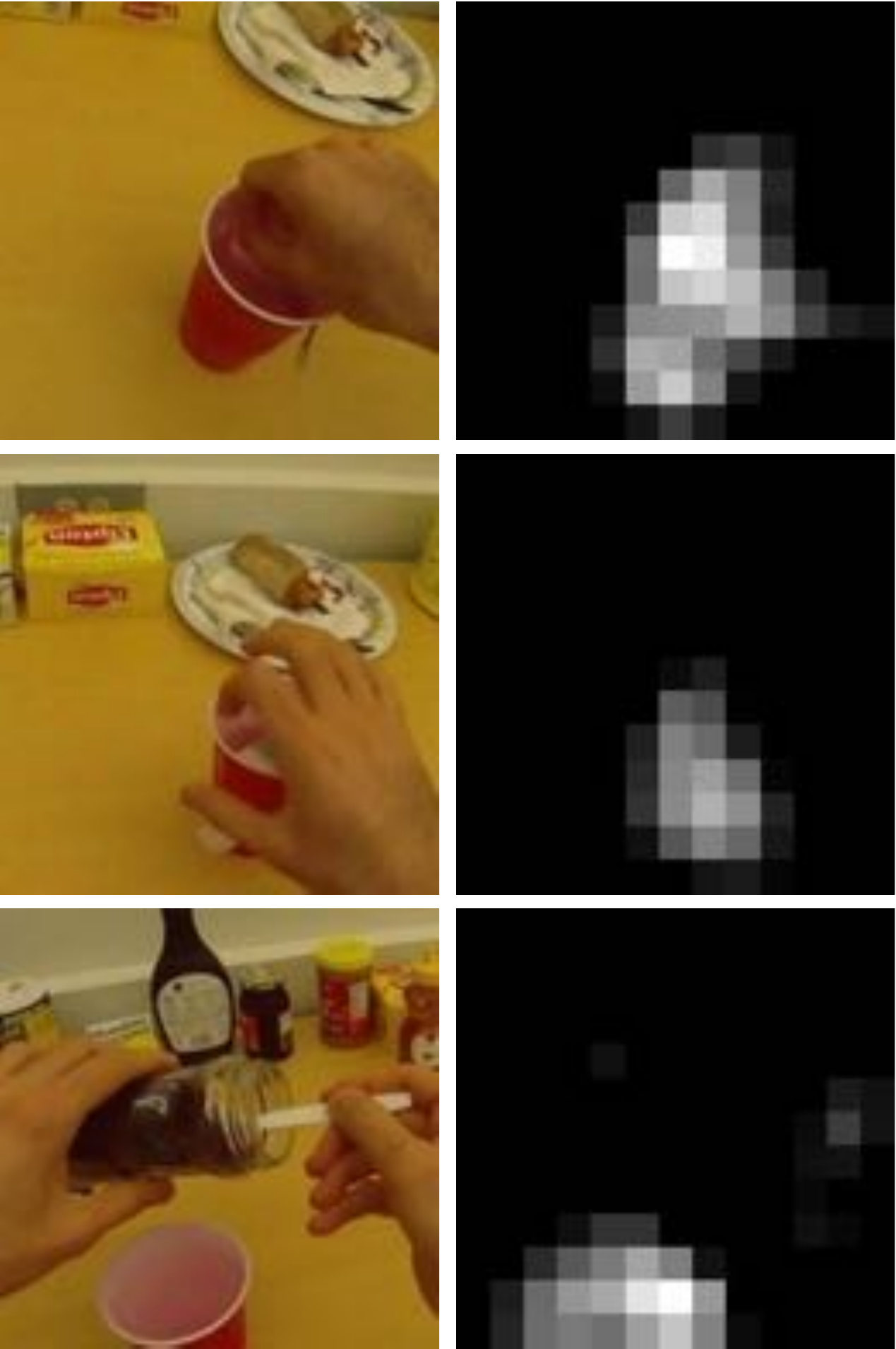}
    \caption{}
    \label{fig:f2}
    \vspace{-2mm}
  \end{subfigure}
  \hspace{2pt}
  \begin{subfigure}[b]{0.11\textwidth}
    \includegraphics[width=\textwidth]{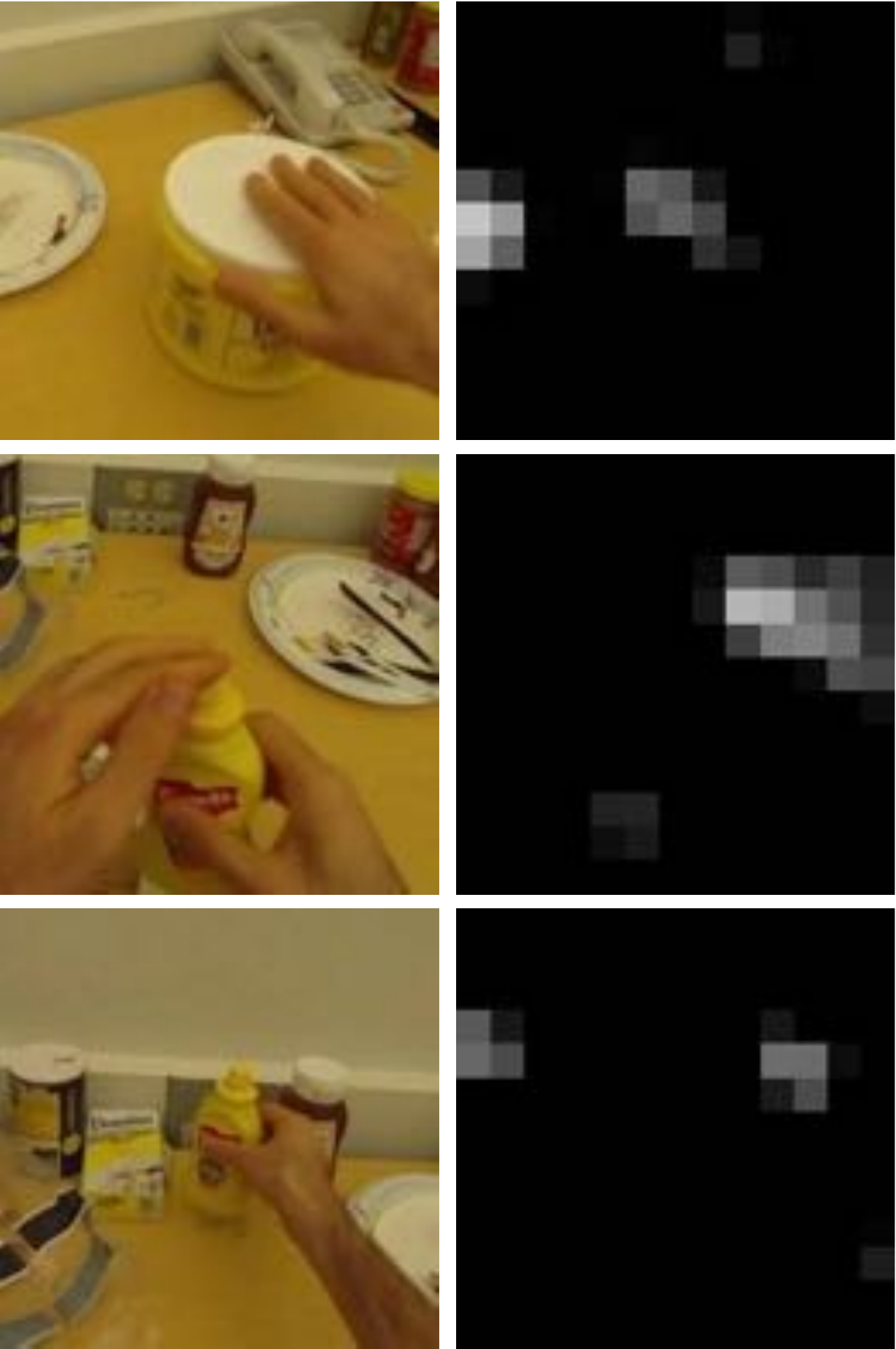}
    \caption{}
    \label{fig:f2}
    \vspace{-2mm}
  \end{subfigure}
\caption{Neuron activation in the $conv5$ layer for test images. Neuron responding to: (a) transparent bottle, (b) edges of container, (c) cups, (d) white round shapes.}
\label{fig-obj-activations}
\vspace{-3mm}
\end{figure}

We visualize the activations of the object recognition network and present two important findings: \textbf{(1) Hands are important for object recognition:} Although the localization network is targeted for object of interest, the cropped image also contains a large portion of hands. We visualize the activations of the $conv5$ layer and find that the $50^{th}$ neuron unit responds particularly strongly to training images with large hand regions as shown in Figure \ref{fig-hand-activations}. We further test the network with test images shown in Figure \ref{fig-hand-activations}. We observe that the strongest activations overlap with hand regions. We therefore conclude that the object recognition network is learning appearance features from hand regions to help recognize objects. When there is no hand in the scene, the localization network will predict no interacting object. Since some of the iterating objects as tea bags and utensils are small, it is extremely challenging to locate them using an traditional object detector. The hands, their shape and their motion can act as a type of proxy small objects. \textbf{(2) Object attributes are important for object recognition:} Figure \ref{fig-obj-activations} shows examples of a particular neuron unit responding to particular object attributes like color, texture and shape. In Figure \ref{fig-obj-activations-b}, we observe that this specific neuron is activated when it observes round objects. 

\vspace*{-1mm}
\subsection{ActionNet performance}\label{sec-action-results}
\vspace*{-1mm}

We first evaluate the ActionNet to recognize actions. In our experiments, we crop and resize input images to $256\times 256$ and calculate optical flow using OpenCV GPU implementation of \cite{zach2007duality}. We clip and normalize the flow values from $[-20, 20]$ to $[0,255]$. We found empirically that $L=10$ optical flow frames generates good performance.

\begin{table}[t]
\renewcommand{\arraystretch}{1.05}
\centering
\footnotesize
\begin{tabular}{l|c|c|c}
\hline
Method \& dataset	    &GTEA(71)	    &Gaze(40)     &Gaze+(44)\\
\hline
Fathi \textit{et al.}\ \cite{fathi2011understanding}  & 47.70          & N/A          & N/A\\
\hline
Motion CNN              & 75.85             & 33.65          & 62.62\\
\hline
Joint training          & \textbf{78.33}          & \textbf{36.27}          & \textbf{65.05}\\
\hline
\end{tabular}
\caption{Average action recognition accuracy. Proposed method performs $30\%$ better than the baseline. Joint training of motion and object networks improves accuracy across all datasets.}
\label{tab-verb-results}
\vspace{-2mm}
\end{table}

Table \ref{tab-verb-results} compares our proposed method with the baseline in \cite{fathi2011understanding}. While our motion network significantly improves the average recognition accuracy, we are also interested in understanding what the network is learning. Our visualization shows two important discoveries: (1) our motion network automatically identifies foreground (objects and hands) motion out of complex background (camera) motion (2) our motion network automatically encodes temporal motion patterns.

\begin{figure}[t]
\vspace{-4mm}
\centering
  \begin{subfigure}[b]{0.271\linewidth}
    \includegraphics[width=\textwidth]{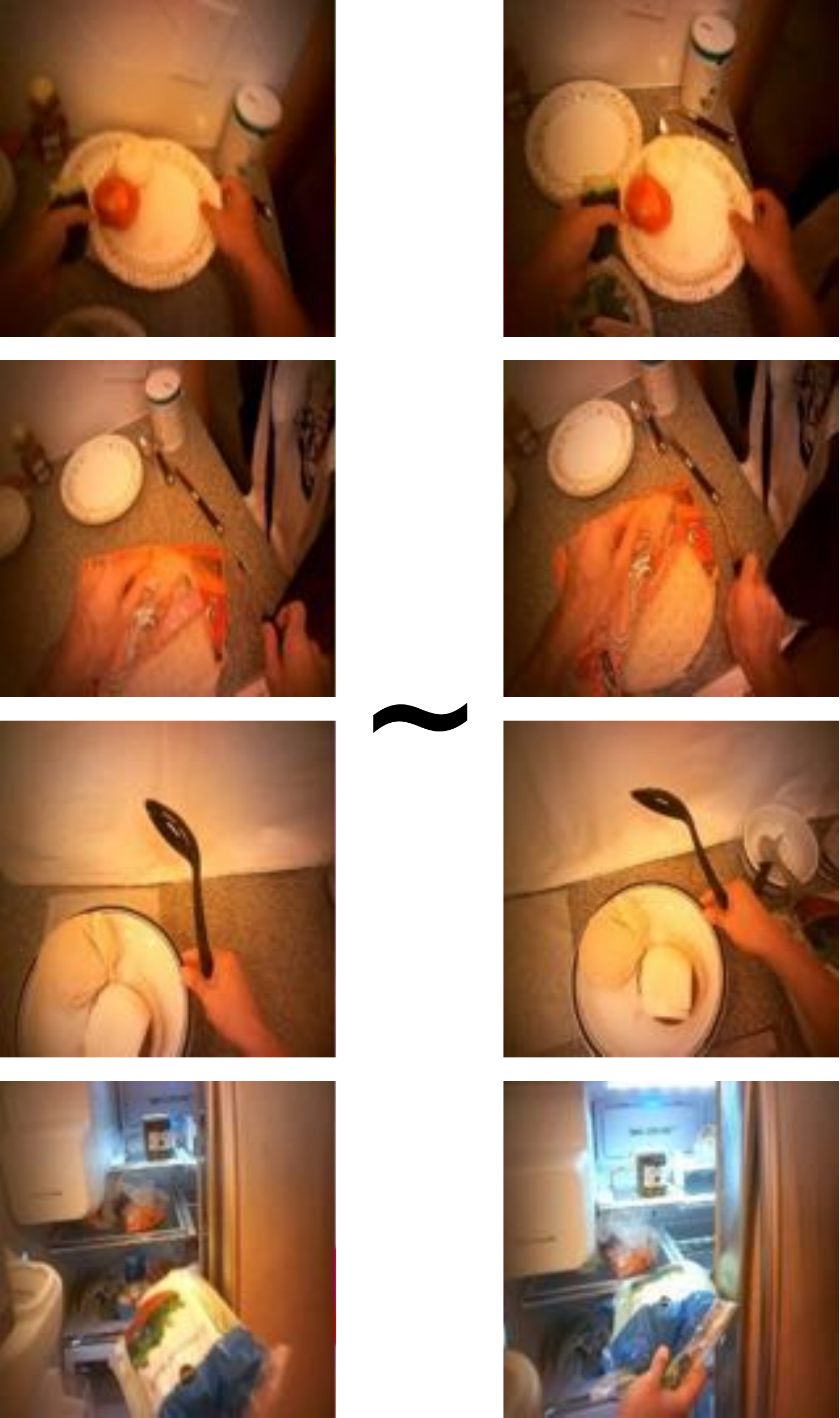}
    \caption{}
    \label{fig:f1}
  \end{subfigure}
  \hspace{2pt}
  \begin{subfigure}[b]{0.271\linewidth}
    \includegraphics[width=\textwidth]{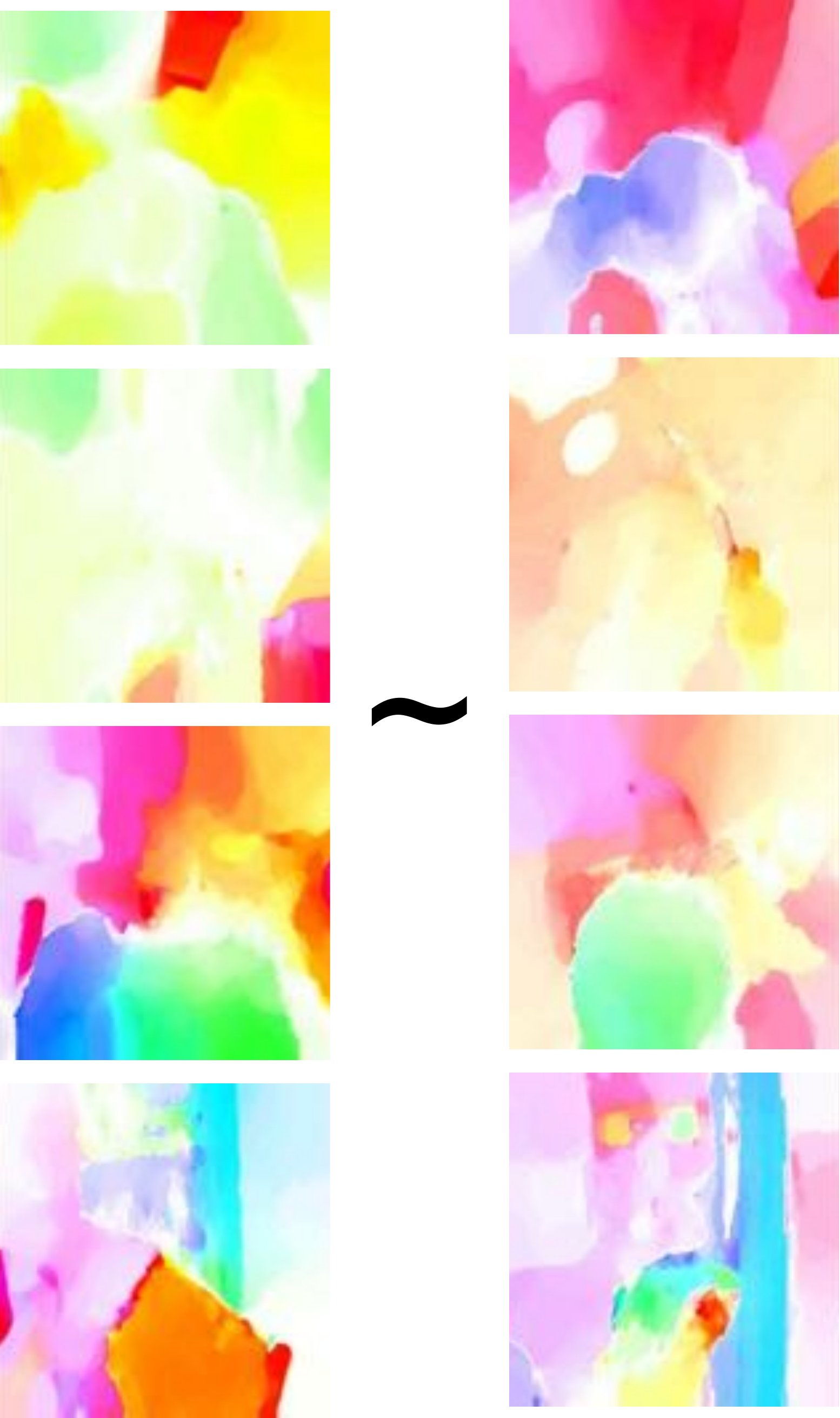}
    \caption{}
    \label{fig:f2}
  \end{subfigure}
  \hspace{2pt}
  \begin{subfigure}[b]{0.109\linewidth}
    \includegraphics[width=\textwidth]{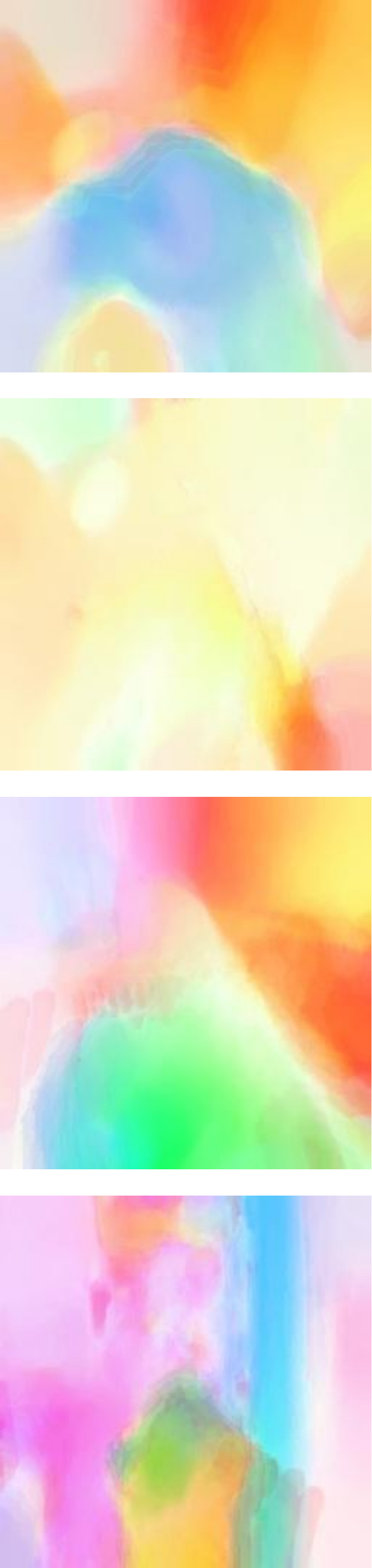}
    \caption{}
    \label{fig:f2}
  \end{subfigure}  
  \hspace{2pt}
  \captionsetup[subfigure]{labelformat=empty}
  \begin{subfigure}[b]{0.049\linewidth}
    \includegraphics[width=\textwidth]{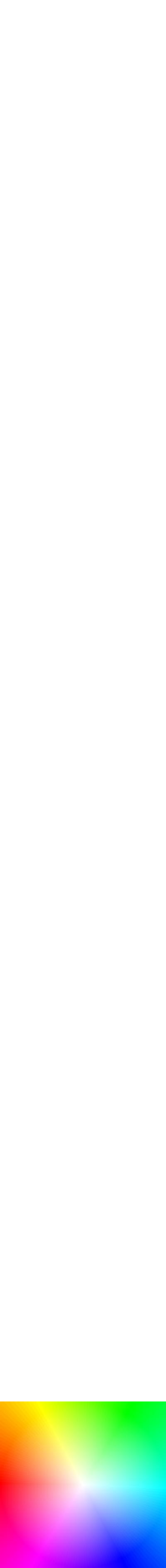}
    \caption{}
    \label{fig:f2}
  \end{subfigure}
\vspace{-2mm}
\caption{Top 4 training sequences with strongest activations for the $346^{\text{th}}$ neuron unit in $conv5$ layer. (a) Start/end image frames, (b) Start/end optical flow images, (c) Average optical flow for each sequence. From top to bottom, ground-truth activity labels are \textit{put\_ cupPlateBowl}, \textit{put\_ knife}, \textit{put\_ cupPlateBowl} and \textit{put\_ lettuce\_ container}.}
\label{fig-top-action-activations}
\vspace{-5mm}
\end{figure}

\begin{figure}[t]
\centering
  \begin{subfigure}[b]{0.222\linewidth}
    \includegraphics[width=\textwidth]{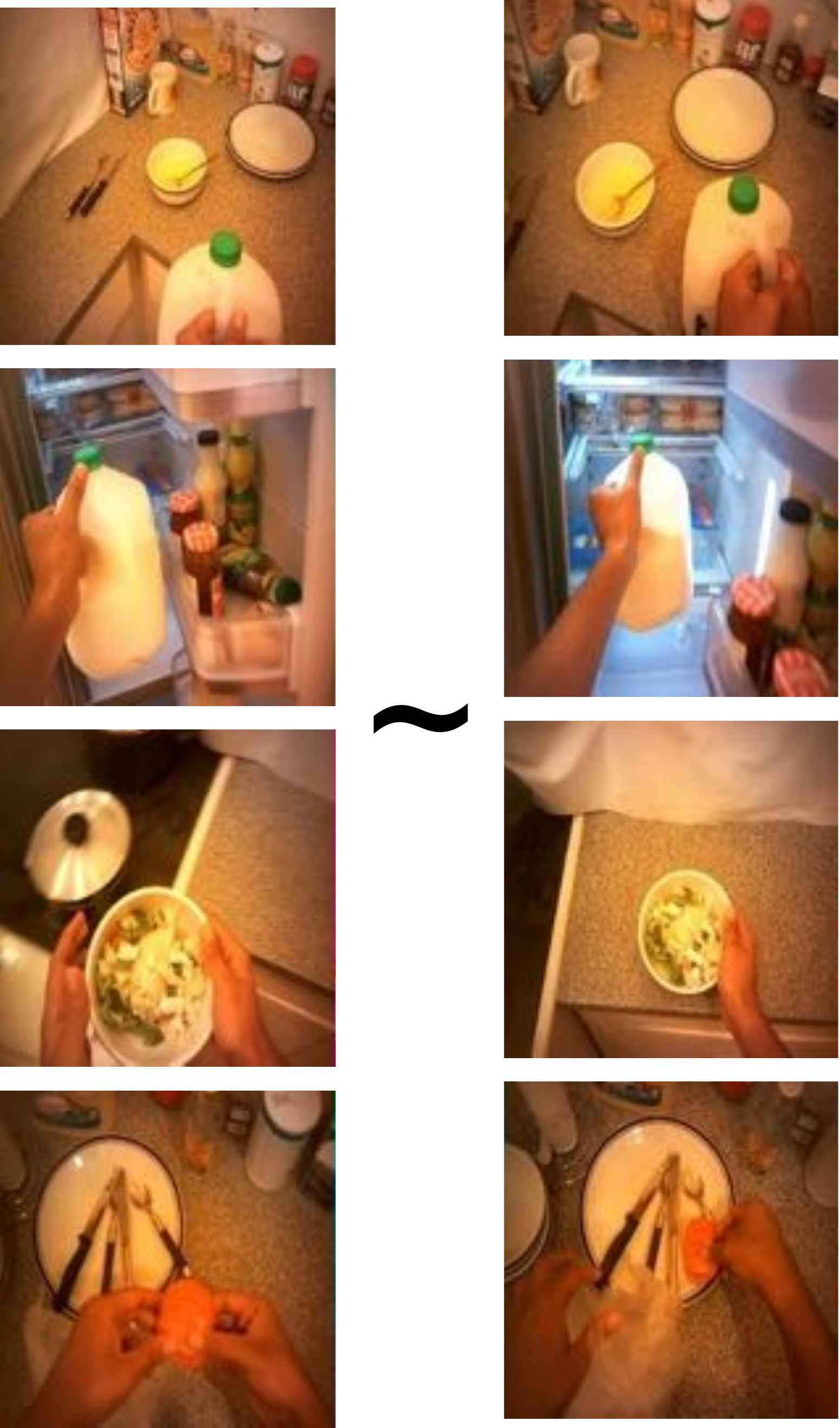}
    \caption{}
    \label{fig:f1}
  \end{subfigure}
  \hspace{2pt}
  \begin{subfigure}[b]{0.222\linewidth}
    \includegraphics[width=\textwidth]{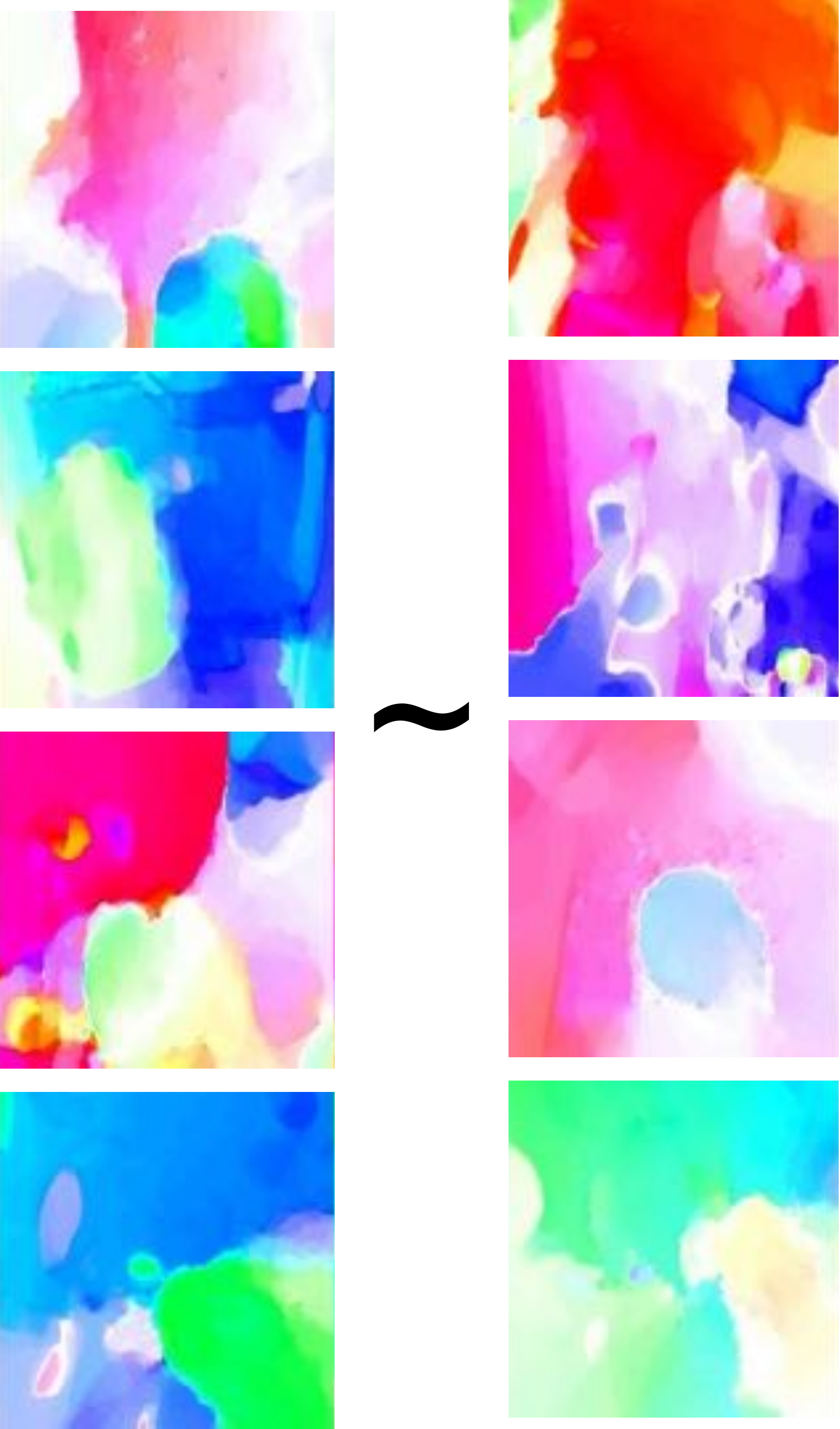}
    \caption{}
    \label{fig:f2}
  \end{subfigure}
  \hspace{2pt}
  \begin{subfigure}[b]{0.09\linewidth}
    \includegraphics[width=\textwidth]{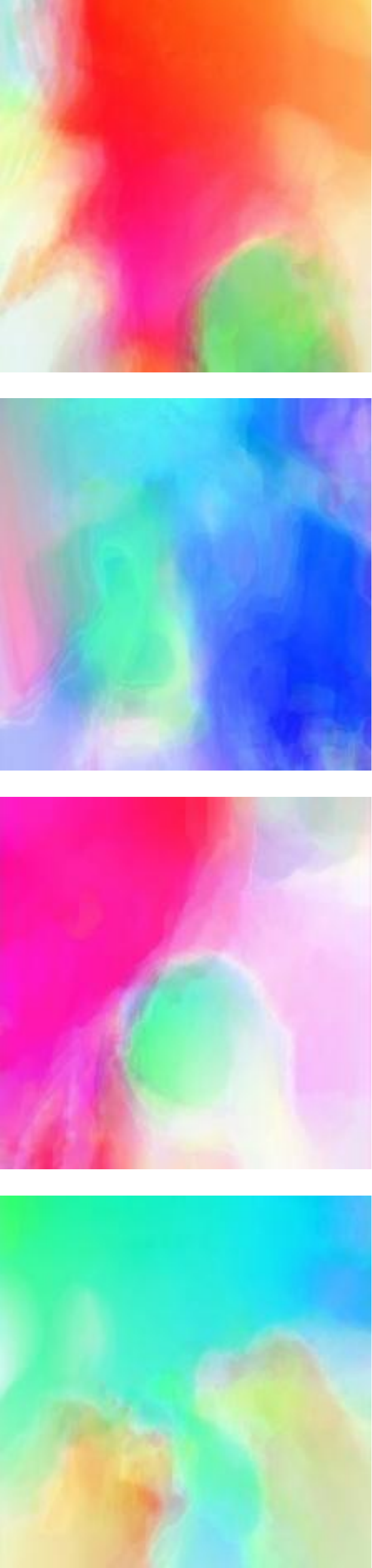}
    \caption{}
    \label{fig:f2}
  \end{subfigure}
    \hspace{2pt}
  \begin{subfigure}[b]{0.09\linewidth}
    \includegraphics[width=\textwidth]{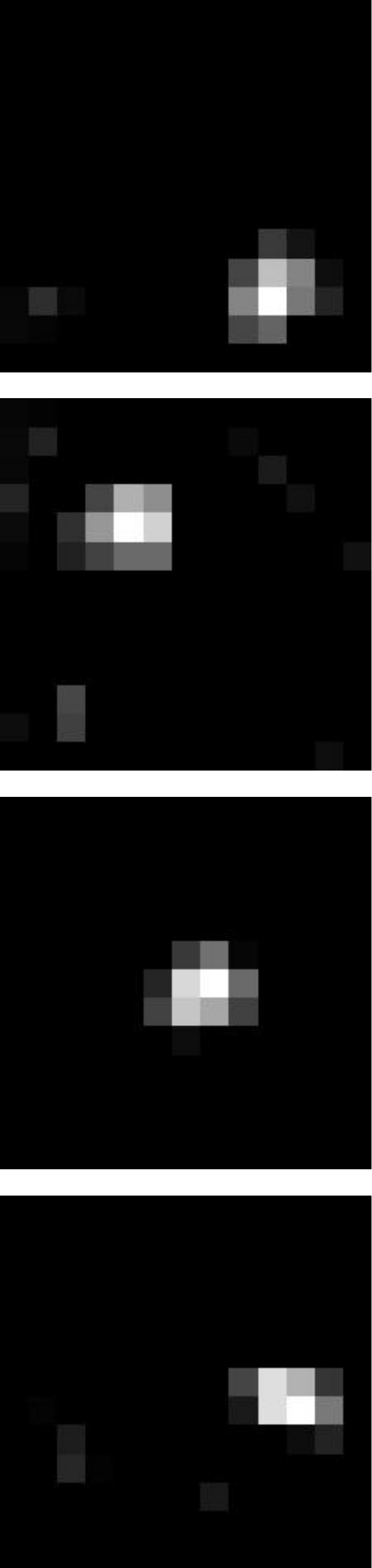}
    \caption{}
    \label{fig:f2}
  \end{subfigure}
    \hspace{2pt}
  \begin{subfigure}[b]{0.09\linewidth}
    \includegraphics[width=\textwidth]{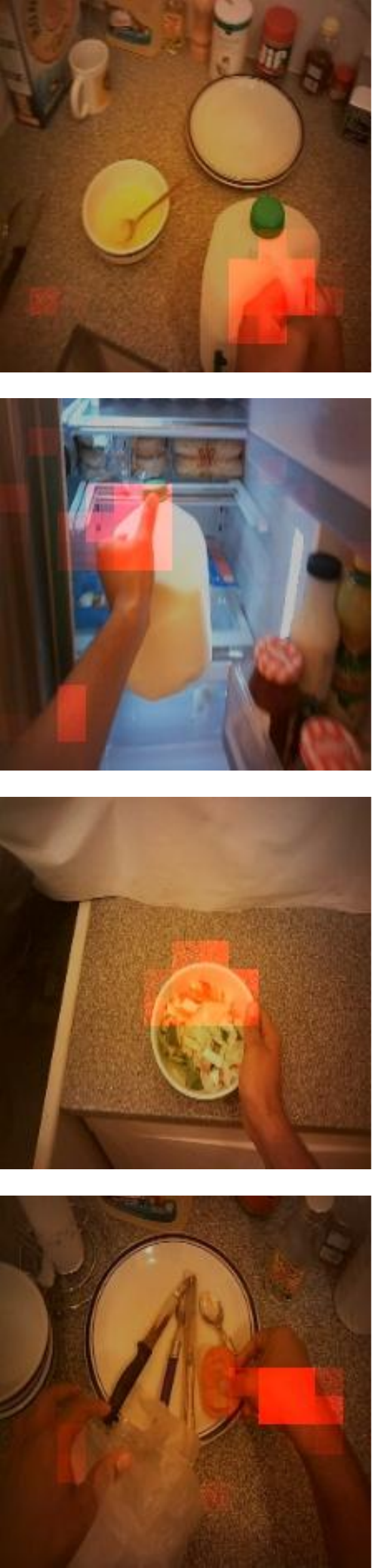}
    \caption{}
    \label{fig-action-activation-overlay-e}
  \end{subfigure}
    \hspace{2pt}
  \begin{subfigure}[b]{0.09\linewidth}
    \includegraphics[width=\textwidth]{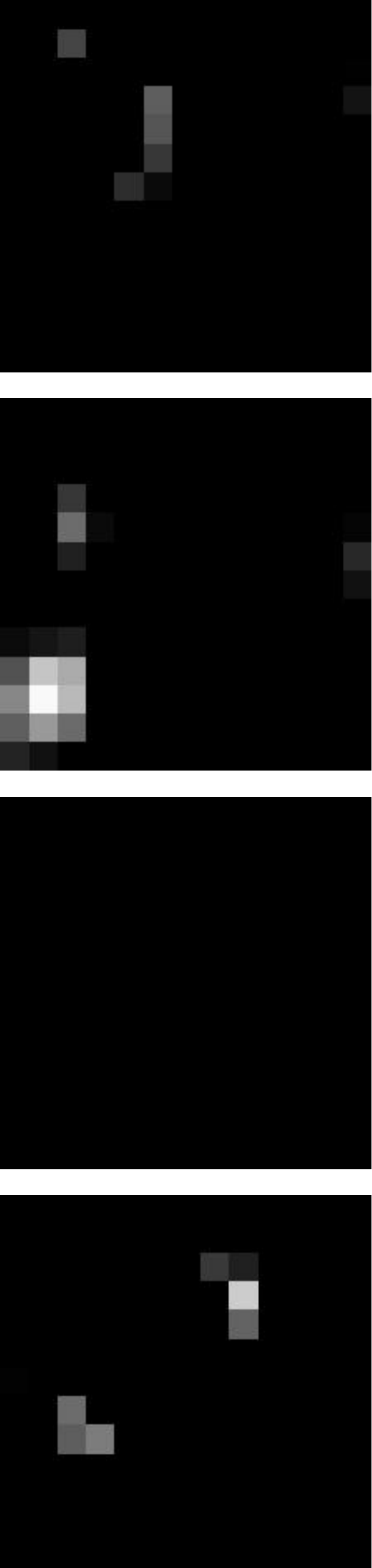}
    \caption{}
    \label{fig-action-activation-test-f}
  \end{subfigure}
\vspace{-2mm}
\caption{4 test sequences and activations of the $346^{\text{th}}$ neuron unit in $conv5$ layer. (a) Start/end image frames, (b) Start/end optical flow images, (c) Average optical flow, (d) $13\times 13$ activation maps of the neuron unit using the optical flow sequence, (e) Overlay of activation map on the end image frame, (f) $13\times 13$ activation maps of the neuron unit using reversed optical flow sequence. From top to bottom, ground-truth activity labels are \textit{put\_ milk\_ container}, \textit{put\_ milk\_ container}, \textit{put\_ cupPlateBowl}, \textit{put\_ tomato\_ cupPlateBowl}.}
\label{fig-action-activation-test}
\vspace{-6mm}
\end{figure}

\begin{table*}[ht]
\newcolumntype{P}[1]{>{\centering\arraybackslash}p{#1}}
\newcolumntype{M}[1]{>{\centering\arraybackslash}m{#1}}
\centering
\footnotesize

\begin{tabular}{c|c|c|c|c|c|c|c}
\hline
&Methods	                        &GTEA(61)$^{\ast}$	    &GTEA(61)     &GTEA(71)    &Gaze(25)$^{\ast}$    &Gaze(40)$^{\ast}$  &Gaze+(44) \\
\hline
\multirow{ 3}{*}{Li \textit{et al.}\cite{li2015delving}} & O+M+E+H   & 61.10          & 59.10          & 59.20          & 53.20         & 35.70          & 60.50 \\
& O+M+E+G   & N/A             & N/A          & N/A            & 60.90          & 39.60          & 60.30 \\
& O+E+H               & 66.80          & 64.00          & 62.10          & 51.10 & 35.10          & 57.40 \\
\hline
\multirow{ 3}{*}{S. \& Z.\cite{simonyan2014two}} & temporal-cnn                   & 34.30          & 30.92              & 30.33          & 38.76          & 22.01      & 44.45 \\
& spatial-cnn                    & 53.77         & 41.13              & 40.16          & 30.84          & 18.46         & 45.97 \\
& temporal+spatial-svm       & 46.51         & 35.69              & 35.81          & 25.94          & 22.18         & 43.23 \\
& temporal+spatial-joint        & 57.64          & 51.58              & 49.65          & 44.29          & 34.70      & 58.77 \\
\hline
\multirow{ 3}{*}{Ours} & object-cnn                         & 60.02          & 56.49        & 50.35          & 47.09          & 35.56          & 46.38 \\
& motion+object-svm                         & 53.01          & 50.45        & 47.07          & 28.42          & 16.00          & 34.75 \\
& motion+object-joint                       & \textbf{75.08}          & \textbf{73.02}         & \textbf{73.24}          & \textbf{62.40}          & \textbf{43.42}          & \textbf{66.40} \\
\hline
\end{tabular}
\caption{Quantitative results for activity recognition. (a) Best results reported from Li \etal \cite{li2015delving}. (b) Two-stream CNN \cite{simonyan2014two} results with single streams, SVM-fusion and joint training. (c) Results from our proposed methods with localized object only, SVM-fusion and joint training. Our proposed joint training model significantly outperforms the two baseline approaches on all datasets. Note that even the network trained using only cropped object images (object-cnn) achieves very promising results thanks to our localization network. ($^{\ast}$: fixed split, O: object, M: motion, E: egocentric, H: hand, G: gaze).}
\label{tab-activity-results}
\vspace{-2mm}
\end{table*}

\begin{figure*}[ht]
\vspace{-4mm}
\centering
\footnotesize
  \begin{subfigure}[b]{0.277\textwidth}
    \includegraphics[width=\textwidth]{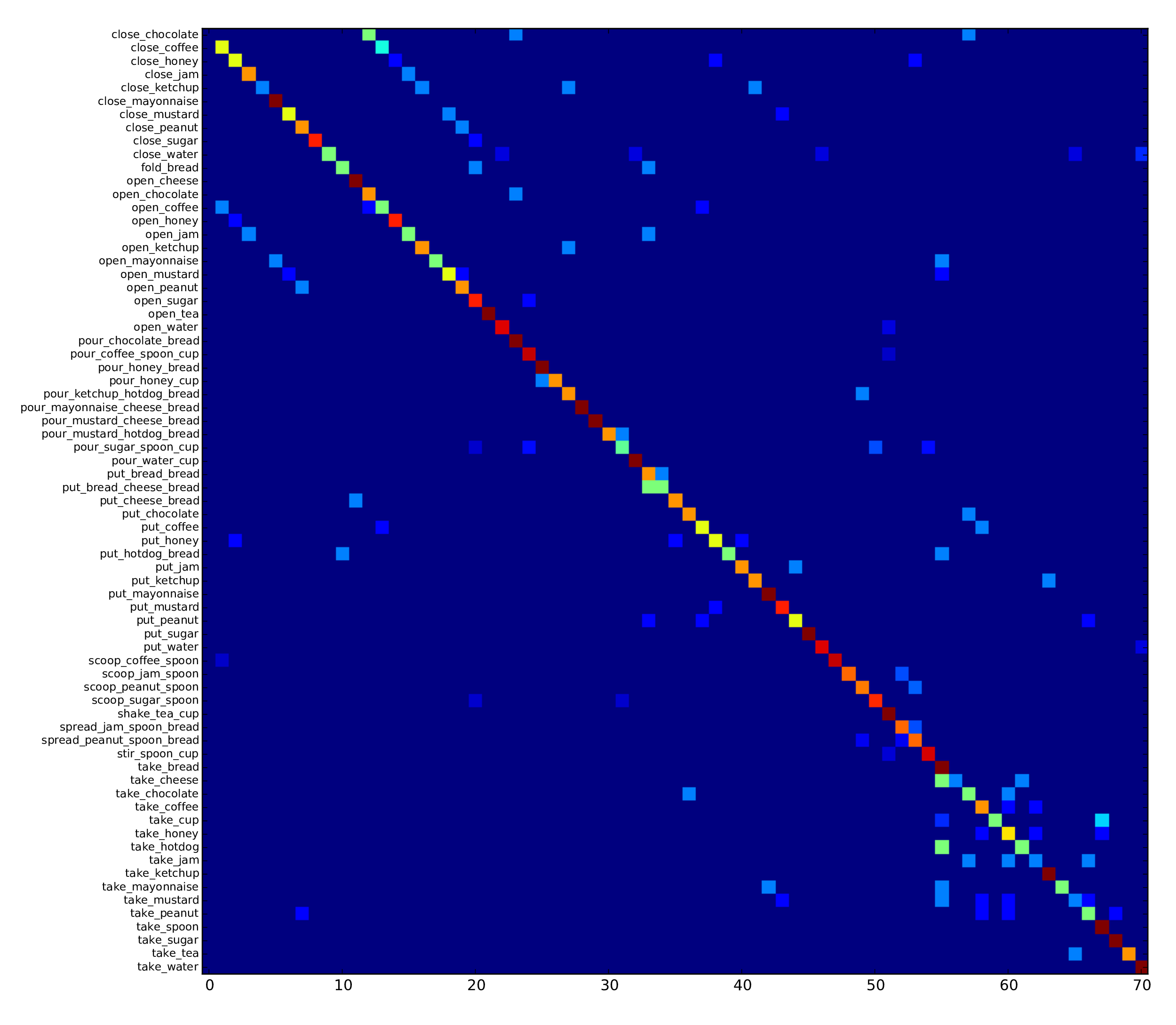}
    \caption{GTEA 71 classes}
    \label{fig:f1}
    \vspace{-2mm}
  \end{subfigure}
  \begin{subfigure}[b]{0.288\textwidth}
    \includegraphics[width=\textwidth]{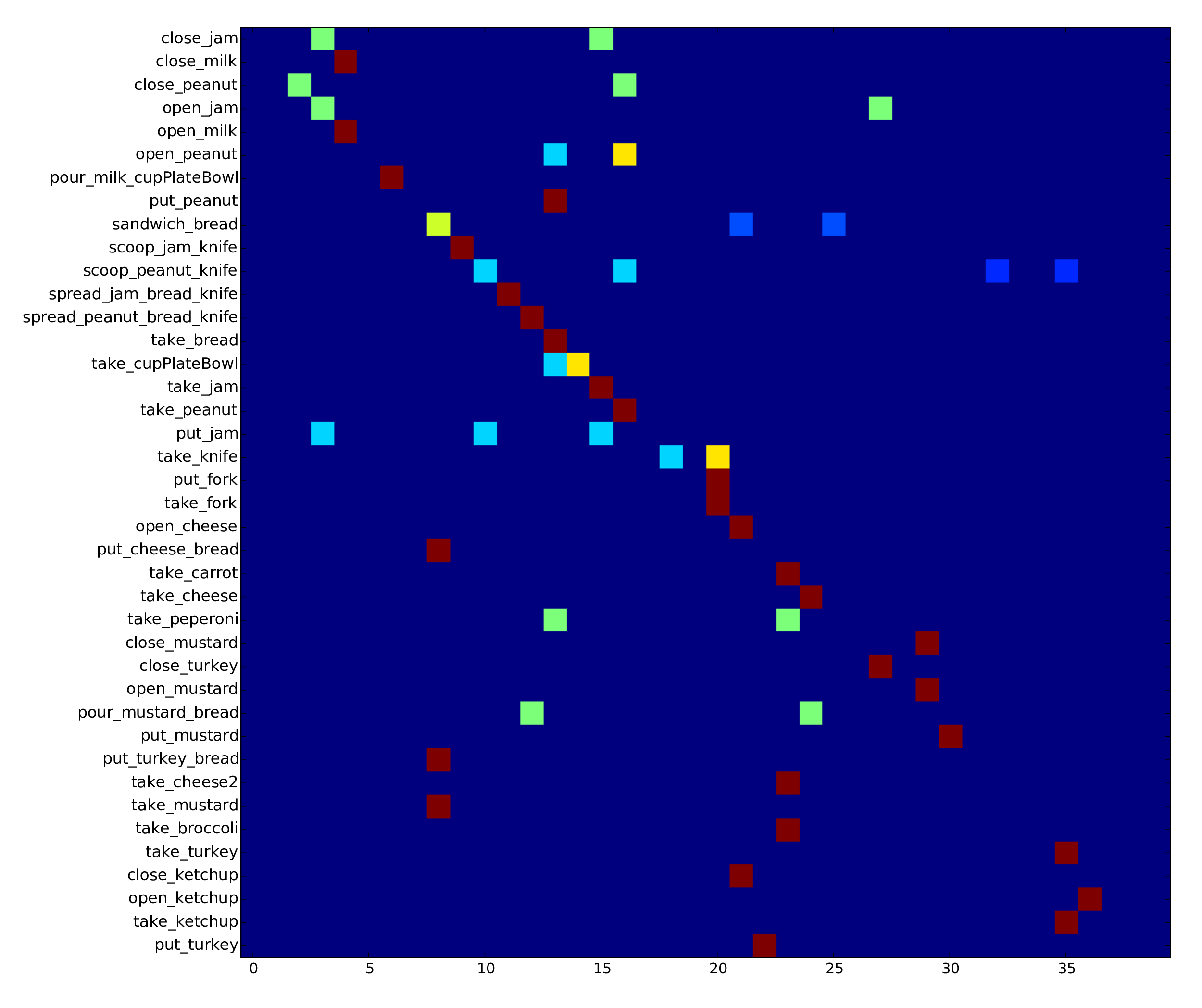}
    \caption{Gaze 40 classes}
    \label{fig:f2}
    \vspace{-2mm}
  \end{subfigure}
  \begin{subfigure}[b]{0.334\textwidth}
    \includegraphics[width=\textwidth]{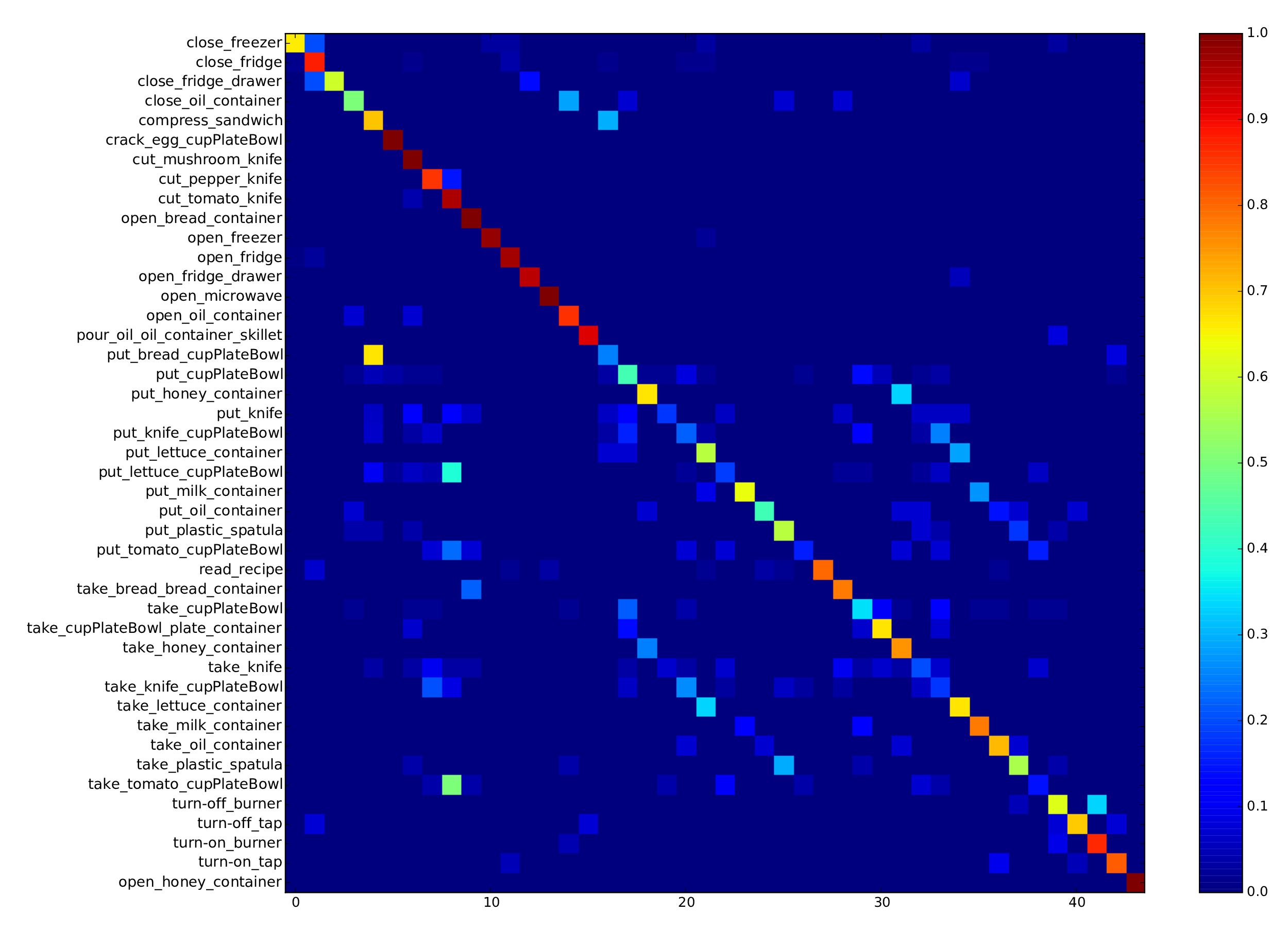}
    \caption{Gaze+ 44 classes}
    \label{fig:f2}
    \vspace{-2mm}
  \end{subfigure}
\caption{Confusion matrices of our proposed method for activity recognition. Improvement on the Gaze dataset is lower due to low video quality and inefficient data. (best view in color)}
\label{fig-obj-activations-confusion-matrix}
\vspace{-6mm}
\end{figure*}

\noindent \textbf{(1) Camera motion compensation is important for action recognition:} As summarized in \cite{li2015delving}, motion compensation is important for ego-centric action understanding as it provides more reliable foreground motion features. Through visualization, we discover that the network is automatically learning to identify foreground objects and hands. Figure \ref{fig-top-action-activations} shows top 4 training sequences that activate a particular neuron unit most strongly in the $conv5$ layer. All these sequences have the same action verb \textit{put} despite the diversity in camera egomotion. This shows that the network automatically learns to ignore background camera motion for this neuron. We further test the network with a few test sequences of \textit{put} actions. The results (in Figure \ref{fig-action-activation-test}) agree with our observation in the following aspects: (1) Activation of the same unit is very strong on all these test \textit{put} actions compared with other actions; (2) The strongest activation location coincides roughly with foreground objects and hands location in Figure \ref{fig-action-activation-overlay-e}.

\noindent \textbf{(2) Temporal motion patterns are important for action recognition:} While instantaneous motion is an important cue for action recognition, it is also crucial to integrate temporal motion information as shown in \cite{li2015delving, ryoo2014pooled, wang2011action, wang2013action}. Figure \ref{fig-top-action-activations} shows that the neuron unit is able to capture the movement of subjects during image sequences. We perform another experiment by reversing the order of the input optical flow images to observe how this neuron responds. Figure \ref{fig-action-activation-test-f} shows the activation maps of the same neuron unit with respect to reversed optical flow frames. The weak responses on suggests that the temporal ordering has been encoded in this neuron. This is reasonable, as actions such as \textit{put} and \textit{take} can only be differentiated be preserving temporal ordering.

\vspace*{-1mm}
\subsection{Activity recognition}\label{sec-activity-results}
\vspace*{-1mm}

We finally evaluate our framework for the task of activity recognition. In this experiment, we concatenate the two fully connected layers from the ActionNet and ObjectNet and add another fully connected layer on top. Then we fine-tune the two streams together with optical flow images, cropped object images and three weighted losses for three tasks. We compare our results with the state-of-the-art method from Li \etal \cite{li2015delving} in Table \ref{tab-activity-results}. We also report results using the two-stream networks from Simonyan and Zisserman \cite{simonyan2014two} without decomposing activity labels. The confusion matrices are shown in Figure \ref{fig-obj-activations-confusion-matrix}. Our proposed method significantly improves the state-of-the-art performance on all datasets. We conclude that this is due to better representations of action and object from the base motion and appearance streams in our framework. We further analyze two main findings from our experiments.

\noindent \textbf{(1) Joint training is effective:} Instead of fixing ActionNet and ObjectNet, and only training stacked layers on top, we jointly train all three networks using three losses as discussed in Section \ref{sec-joint-training}. This avoids over-fitting the newly added top layers and leads to a joint representation of activities with actions and objects. In our experiments, we set $w_{\text{action}}=0.2$, $w_{\text{object}}=0.2$ and $w_{\text{activity}}=1.0$. We set the activity loss weight higher for faster convergence of activity recognition. We also compare joint training with SVM fusion of two networks in Table \ref{tab-activity-results}. Joint training boosts the performance consistently by $27\%$ over all datasets.  

\noindent \textbf{(2) Object localization is crucial:} We seek to understand the importance of localizing objects by training a network using cropped object images and activity labels. We compare three networks for activity recognition with best results reported in \cite{li2015delving}: (1) motion-cnn (temporal-cnn) using optical flow images and activity labels (2) spatial-cnn using raw images and activity labels (3) object-cnn using cropped object images and activity labels. The performance is lower than \cite{li2015delving} on three networks as shown in Table \ref{tab-activity-results} because we are not using any motion or temporal information. However, the performance of object-cnn is surprisingly close, only $9.6\%$ lower ($25.5\%$ lower with motion-cnn, $20.6\%$ lower with spatial-cnn) on average. We conclude that localizing the key object of interest is crucial for egocentric activity understanding.

\vspace*{-1mm}
\section{Conclusion}
\vspace*{-1mm}

We have developed a twin stream CNN network architecture to integrate features that characterize ego-centric activities. We demonstrated how our proposed network jointly learns to recognize actions, objects and activities. We evaluated our model on three public datasets and it significantly outperformed the state-of-the-art methods. We further analyzed what the networks were learning. Our visualizations showed that the networks learned important cues like hand appearance, object attribute, local hand motion and global ego-motion as designed. We believe this will help advance the field of ego-centric activity analysis.

\section*{Acknowledgement}
This research was funded in part by a grant from the Pennsylvania Department of Health's Commonwealth Universal Research Enhancement Program and CREST, JST.

{\small
\bibliographystyle{ieee}
\bibliography{egbib}
}

\end{document}